# captions

# SEEING VOICES
## Generating A-Roll Video from Audio with Mirage

Aditi Sundararaman    Amogh Adishesha    Andrew Jaegle    Dan Bigioi    Hyoung-Kyu Song
Jon Kyl    Justin Mao    Kevin Lan    Mojtaba Komeili    ShahRukh Athar    Sheila Babayan
Stanislau Beliasau    William Buchwalter*

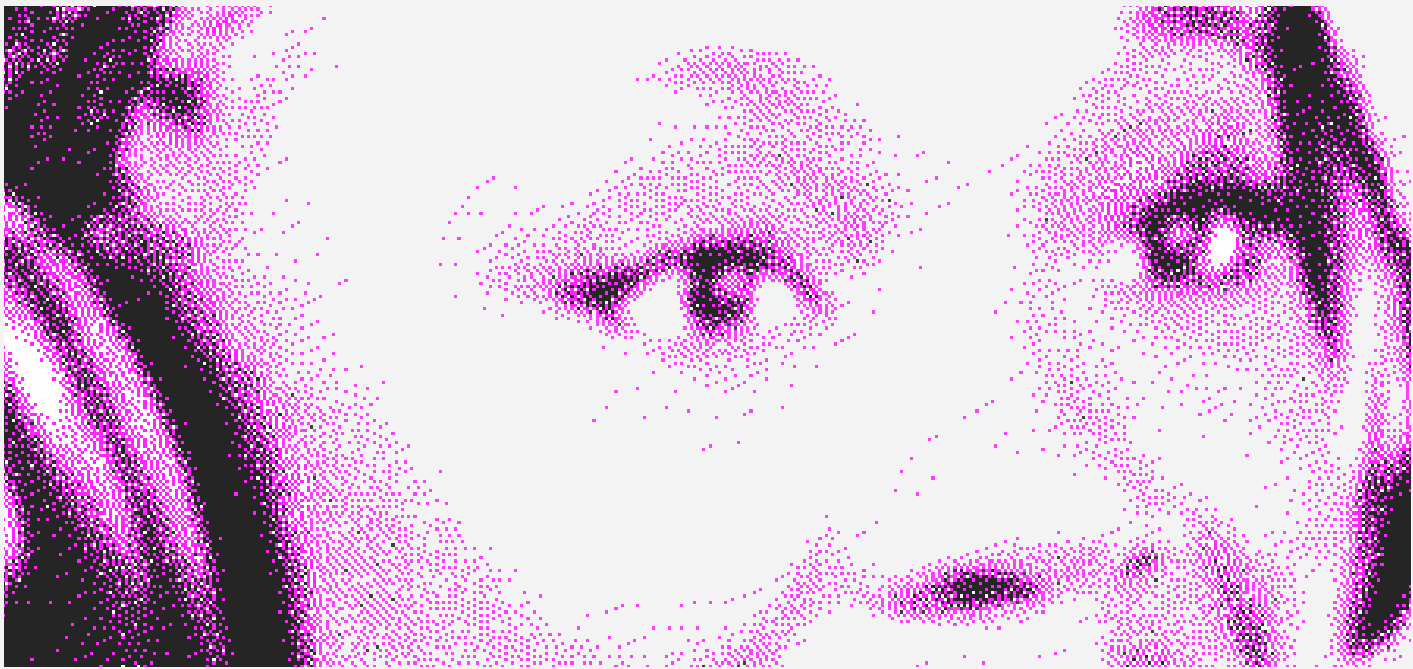

## ABSTRACT

From professional filmmaking to user-generated content, creators and consumers have long recognized that the power of video depends on the harmonious integration of what we hear (the video's audio track) with what we see (the video's image sequence). Current approaches to video generation either ignore sound to focus on general-purpose but silent image sequence generation or address both visual and audio elements but focus on restricted application domains such as re-dubbing. We introduce Mirage, an audio-to-video foundation model that excels at generating realistic, expressive output imagery from scratch given an audio input. When integrated with existing methods for speech synthesis (text-to-speech, or TTS), Mirage results in compelling multimodal video. When trained on audiovideo footage of people talking (A-roll) and conditioned on audio containing speech, Mirage generates video of people delivering a believable interpretation of the performance implicit in input audio. Our central technical contribution is a unified method for training self-attention-based audio-to-video generation models, either from scratch or given existing weights. This methodology allows Mirage to retain generality as an approach to audio-to-video generation while producing outputs of superior subjective quality to methods that incorporate audio-specific architectures or loss components specific to people, speech, or details of how images or audio are captured. We encourage readers to watch and listen to the results of Mirage at [mirage.app/research/seeing-voices](mirage.app/research/seeing-voices) and to try Mirage for themselves at: [https://mirage.app](https://mirage.app).

*Authors listed alphabetically by first name.*
*\*Work done while at Captions*

> Films are 50 percent visual and 50 percent sound.
> Sometimes sound even overplays the visual.
> — David Lynch, 1946-2025

# INTRODUCTION

Since the invention and widespread adoption of synchronized sound in the first thirty years of the 20th century, audio-visual video has been one of primary form of information transmission in every medium it has touched. Cinema grew from an imagery-only format to an audio-visual format in the late 1920s with the advent of the "talkies" (Thompson and Bordwell, 2009). Television has been audiovisual since its earliest forms, and even the initial proof-of-concept broadcasts synchronized audio with video (Edgerton, 2009). The Internet of the late 2010s to mid-2020s has been increasingly dominated by platforms that foreground user-generated and professionally created audio-visual content.

Imagery and audio provide complementary streams of information, especially in the case of "A-roll" footage. A-roll refers to the primary, narrative-advancing portions of a video, such as scenes where actors deliver lines onscreen. In material of this kind, nuances both coarse and subtle arrive in the words we hear, how they're delivered, and the auditory scene in which they occur: the audio conveys the invisible. The video conveys details that can only be seen about the relationship between the speaker and objects and others in the scene. Audio and video work together in lock-step to convey what is happening and its physical and emotional characteristics: we see and hear so that we may know and feel.

And yet general-purpose video generation (one of the most active areas of research in AI in the mid-2020s) is bifurcated between general-purpose techniques adapted primarily for image sequence generation (*silent* video) and techniques for non-silent video generation that make domain-specific assumptions to allow for audio-visual generation or synchronization in certain application areas of interest, such as lip dubbing or sound FX generation. This has resulted in a landscape of general-purpose models that generate compelling footage without audio, and specialized models that animate existing images or image sequences to match audio.

In this report, we introduce the Mirage model, which is designed to generate video from audio, even without a reference image or text description of the desired imagery. Unlike previous approaches in this space, Mirage avoids domain-specific assumptions in its design, allowing it to adapt to the audio-visual properties of its training data without requiring architecture or training loss changes, and allowing it to use a single recipe to easily incorporate conditioning signals from additional modalities. Mirage uses a Diffusion Transformer (DiT) architecture and relies solely on asymmetric self-attention (Genmo Team, 2024) to mix information between audio and image sequence modalities, preserving the homogeneous modular design of the base architecture. We find that the self-attention layers can be trained to balance attention to video, audio, text, and other modalities through a simple warm-up and stitching recipe that works either when training from scratch or when finetuning from a silent video model. This same architecture design and training procedure extends to other modalities, like image reference conditioning, allowing us to easily extend Mirage to handle new conditioning signals.

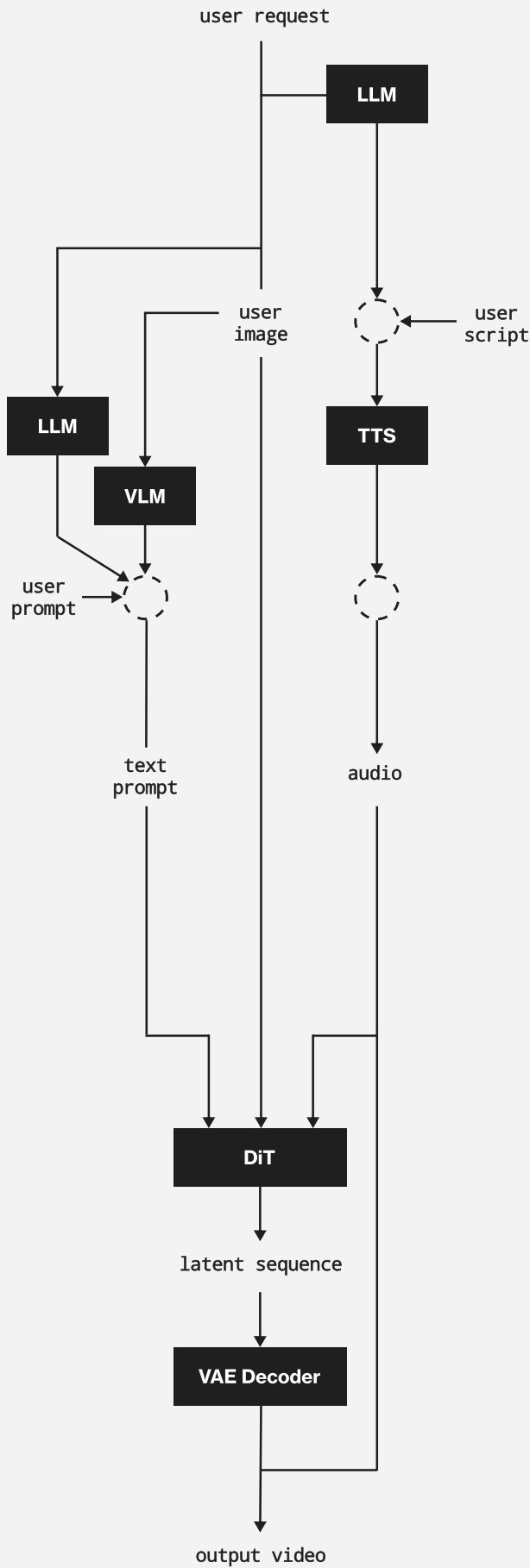

When trained on data that is specialized for the task of A-roll generation (a special case of which is talking person video generation), Mirage produces outputs that are highly expressive and realistic, while also matching the content of the audio stream, reflecting features of speaker appearance, lip/face/body-gesture timing and intensity, and other properties of the input audio. Mirage reveals surprising aspects of A-roll scenes, identifying when and how speaking subjects move in relationship to a static microphone, when cuts and splices are made in the source video, and the intuitive physics of the acoustic world like movement relative to microphone and properties of the recording space. When trained at large-scale, Mirage is able to visually depict physical attributes leading to outputs that have high appearance-timbre congruence and avoiding the uncanny valley effect produced by lipdubbed images with mismatched voices.

● Figure 1

Mirage system diagram. The system enables users to generate A-roll from audio, text, or reference images (either alone or in combination). Given this input, the Mirage Diffusion Transformer (DiT) model generates image sequences for 4 or 8 second videos. If conditioning audio is unavailable, full A-roll can be generated using text requests through the use of a large language model (LLM) that converts user requests into scripts, an LLM that converts user requests into text prompts, a text-to-speech (TTS) model that converts scripts into voiced audio. Similarly the system allows users to generate video based on a reference image (for example depicting a person in some setting with some photographic conditions) by uploading an image and optionally describing the image in text with a vision-language model (VLM). The output of the DiT is decoded to viewable video frames using a variational autoencoder (VAE) decoder, and merged with the input or generated audio to produce the final output video.

## 1.1. TECHNICAL CONTRIBUTIONS

We present a unified method for training self-attention-based audio-to-video generation models, either from scratch or given existing weights, without the need for audio-to-video cross-attention blocks or other specialized cross-modality modules. We adapt the flow-based model pipeline with minimal tweaks to generate flexible, expressive video that precisely matches input audio. We build on the Mochi variant of the Diffusion Transformer (DiT) architecture [(Genmo Team, 2024)](#). We do so by developing:

```
A unified system for A-roll generation from audio and multimodal
conditioning.
```

```
Extensions of Mochi's asymmetric attention and learned rotary
position encoding (RoPE) strategy to multimodal-conditioned video
generation.
```

```
A robust data preprocessing and filtering pipeline for building our
A-roll generation dataset.
```

This approach allows us to train Mirage for expressive, photorealistic A-roll generation and to generate video unconditionally or conditioned on audio, text, or reference images together or separately. See [Fig. 1](#) for an overview of the Mirage system as a whole. As of the date of publication, Mirage is integrated into several Captions products. We encourage interested readers to explore these product surfaces to see and hear what Mirage can do.

# RELATED WORK

## 2.1. OVERVIEW

Mirage builds on prior work in multimodal-conditioned video generation and its various branches such as text-to-video, image-to-video, and audio-to-video synthesis. As an approach to A-roll generation, Mirage builds on and contributes to application domains focused on generating expressive videos of people. We begin with a review of the broad field of video generation before surveying the specific bodies of work around co-speech gesture generation (generating body motions to match speech), talking head synthesis, and audio representation learning. Mirage synthesizes this work to produce output A-roll videos with unprecedented levels of realism and performative expressivity.

## 2.2. VIDEO GENERATION

Over the past decade, video generation has developed dramatically from early work using recurrent and convolutional networks to model simple, low-resolution, academic image data to contemporary work using Transformer-based architectures to capture the dynamics of high-resolution real-world data with multimodal context. Early work used recurrent networks such as Long Short-Term Memory units [(LSTMs, Hochreiter and Schmidhuber 1997)](#) to learn representations of video sequences using simple reconstruction objectives [(Jaegle et al., 2018](#); [Srivastava et al., 2015)](#). Subsequent work developed temporal and recurrent convolutional architectures and incorporated more sophisticated generative modeling approaches such as Variational Autoencoders (VAEs) ([Babaeizadeh et al., 2018](#); [Denton and Fergus, 2018](#)) and generative adversarial networks (GANs) ([Brooks et al., 2022](#); [Clark et al., 2019](#); [Goodfellow et al., 2014](#); [Mathieu et al., 2016](#); [Tulyakov et al., 2018](#)). This line of work produced compelling results on settings with simplified visual dynamics, but typically struggled to produce high-quality outputs over longer time spans or in complex real-world settings.

In the early-to-mid 2020s, spatiotemporal U-Net-based architectures ([Ho et al., 2020](#); [Nash et al., 2023](#); [Ronneberger et al., 2015](#)) and codec-based strategies (notably VAE-based neural codecs, e.g. [An et al. 2023](#); [Blattmann et al. 2023](#); [OpenAI 2024](#); [Wang et al. 2025, 2024](#)) facilitated the expressive, high-dimensional computation needed to handle larger images, longer sequences, and less constrained real-world visual dynamics. Strategies built around scalable Transformer-based architectures ([Jaegle et al., 2022](#); [Peebles and Xie, 2023](#); [Vaswani et al., 2017](#)) and general-purpose modeling techniques such as autoregression and diffusion emerged as powerful general-purpose techniques for generating many modalities, including video (e.g. Hawthorne et al. 2022; Ho et al. 2022; Kalchbrenner et al. 2016; Weng 2024; Yan et al. 2021). Denoising diffusion in particular (Ho et al., 2020; Sohl-Dickstein et al., 2015) revolutionized the field, building out a set of techniques for incrementally removing additive noise from training data to produce high-quality samples from a training set. More recently, the flow-matching or rectified flow paradigm for diffusion-like iterative inference of transition paths through high-dimensional space (Lipman et al., 2023; Liu et al., 2023b) has led to impressive results in video generation. Flow-matching directly models trajectories through probability space, typically in a deterministic manner via ordinary differential equations (ODEs) to transform a simple noise distribution (typically a standard Gaussian) into the target distribution operationalized by a large-scale training dataset.

Many new strategies for multimodal conditioning have developed in parallel. Following the breakthrough performance of text-conditioned diffusion models for static image generation (Dhariwal and Nichol, 2021), researchers have extended these techniques to video synthesis (Ho et al., 2022). Text can serve as a prior for understanding structural layout as well as temporal changes (Singer et al., 2022). Explicit motion priors or motion modeling can help in temporally consistent video generations (Guo et al., 2024; Wu et al., 2023; Xu et al., 2024). For increased precision and user-friendliness in control interfaces, other approaches use image conditioning in addition to text (Girdhar et al., 2024; Xing et al., 2024; Zhang et al., 2023). Building upon these advances in text-to-video, image-to-video, and motion prior techniques, some groups have explored alternative generative frameworks to address the computational challenges and quality limitations of diffusion models (Kong et al., 2024). Other work has explored using explicit 3D annotations such as point clouds, segmentation masks and landmarks, or parametric body models to constrain and control video generation (Chen et al., 2025; Liu et al., 2025; Pan et al., 2019; Yariv et al., 2025; Zhu et al., 2024). Mirage uses audio, text, and images to control and constrain video generation, as these three signals frequently appear alongside A-roll image sequences naturally, co-vary with image dynamics in intuitive and informative ways, and can be extracted at high quality.

As methods for video generation have improved, their areas of application has expanded. Early work typically focused on video generation as a proxy unsupervised task to learn representations useful for downstream tasks such as classification or action recognition (Srivastava et al., 2015; Vondrick et al., 2016). As the quality of models and the reliability of conditioning strategies improved, video generation models have found a wide range of applications including domains where accurate world modeling is beneficial like control and planning (e.g. Ha and Schmidhuber 2018; Hafner et al. 2024; Oh et al. 2015; Rybkin et al. 2019) or physical simulation (e.g. Botev et al. 2021; Toth et al. 2020). As perceptual realism and expressive quality has improved, the outputs of video models are becoming useful artifacts in their own right, whether for creative filmmaking, advertisement, or entertainment. Mirage is designed for the generation of perceptually appealing videos intended for broad consumption.

## 2.3. SPEECH-DRIVEN HUMAN ANIMATION

As A-roll focuses on people performing through audio and video, Mirage shares commonalities with the field of speech-driven human animation. The goal of speech-driven human animation is to generate expressive talking or gesturing humans synchronized with speech or other audio input.

Success here requires synthesizing temporally coherent and visually realistic human motions (e.g. facial expressions, lip movements, body gestures) conditioned on speech signals. We briefly review work in this field focused on talking head and generation and co-speech gesture generation (i.e. synthesizing gestures from human audio) and their relationship to Mirage.

### 2.3.1 TALKING HEAD GENERATION

Building on early rule-based systems (Cohen et al., 2002), the field of talking head generation underwent a paradigm shift with the work of Karras et al. (2017) and Taylor et al. (2017), who pioneered data-driven methods to map raw audio signals to facial motion. The field has seen an extensive body of work exploring diverse approaches to this challenge, ranging from end-to-end generative frameworks leveraging adversarial training (e.g. Chen et al. 2018; Eskimez et al. 2020; Mittal and Wang 2020; Prajwal et al. 2020) to modular pipelines (e.g. Cudeiro et al. 2019a; Lahiri et al. 2021; Zhang et al. 2021a,b; Zhou et al. 2020) that decompose the problem hierarchically: first generating motion through facial landmarks or parametric face models (e.g. FLAME, Li et al. 2017), then subsequently applying neural rendering techniques to synthesize photorealistic outputs (Athar et al., 2022; Gafni et al., 2021; Grassal et al., 2022; Zheng et al., 2022). More recently, diffusion models have shown tremendous success on this task, by selectively denoising partially masked views or close-up images of faces (Bigioi et al., 2024; Cui et al., 2025; Lin et al., 2025; Shen et al., 2023; Stypułkowski et al., 2024; Tian et al., 2024; Xu et al., 2024). Mirage unlocks unprecedented performance on these tasks by leaning heavily on the generality and architectural simplicity enabled by large-scale iterative full-video denoising for generative modeling coupled with rich, multimodal data.

### 2.3.2 CO-SPEECH GESTURE GENERATION

Speech-driven human animation depends on the synthesis of gestures that match speech. Successfully performing this task depends on the ability to model the one-to-many property of speech and gestures. For example, a verbal "okay" may be accompanied by a thumbs-up gesture, a head nod, a bright smile, or a neutral face, and a good model must capture this. Early animation systems relied on handcrafted rules to synchronize predefined motions with speech rhythm or assigned probabilities to gesture data (Cassell et al., 2001; Kopp and Wachsmuth, 2004; Pelachaud and Bilvi, 2003) and sampling motions from the data given a speech input rather than synthesizing from scratch (Bergmann and Kopp, 2009; Kipp, 2005; Neff et al., 2008). Early deep-learning based methods demonstrated the viability of learning gesture directly dynamics from audio-visual datasets (Chiu et al., 2015; Ferstl and McDonnell, 2018; Hasegawa et al., 2018; Yoon et al., 2019). Early deterministic approaches, often reliant on recurrent architectures Ferstl and McDonnell (2018); Hasegawa et al. (2018), produced motions that lacked natural variability, frequently exhibiting mechanical rigidity due to their inability to account for the inherent one-to-many mapping between speech and plausible gestures. To address this one-to-many property, the field increasingly embraced probabilistic frameworks Alexanderson et al. (2020); Ghorbani et al. (2022); Li et al. (2021); Taylor et al. (2021), which explicitly model uncertainty in gesture prediction, enabling diverse and contextually adaptive motions. Recent work leverages diffusion models for further refinement, producing gestures with unprecedented naturalness and expressivity (Alexanderson et al., 2023; Chhatre et al., 2024). Each of the methods discussed above primarily animate avatars by generating the appropriate gestures and movements in response to speech input, but they still require rendering to produce the final video output. Mirage learns the joint statistics of audio and video in a unified latent-space generative model, allowing samples to capture fluid performances directly (see Sec. 5).

Mirage makes modeling choices that allow it to apply to arbitrary A-roll generation. Mirage is trained on a large, diverse range of multimodal A-roll video data, covering human actions, object interactions, dynamic scenes, and more. By unifying tasks like talking head generation, gesture synthesis, and full-body motion prediction, Mirage breaks the boundaries between specialized avatar models, human-specific generation tasks, and open-domain video synthesis.

## 2.4. AUDIO REPRESENTATION LEARNING

Choosing an appropriate audio representation is crucial for training an end-to-end audio-to-video generator that produces realistic, engaging videos. Features like Mel-frequency cepstral coefficients (MFCCs) or Mel spectrograms have been used for tasks such as talking head generation (Zhong et al., 2023; Zhou et al., 2020) and co-speech gesture synthesis (Alexanderson et al., 2023). However, these features are often overly general for the downstream goal of capturing the subtle nuances spread between speech and gesture, as they focus on capturing extraneous audio information that doesn't directly aid in generating expressive movements. This can limit the scalability and adaptability of models, making them less effective for real-world applications that require dynamic, contextually appropriate responses.

Cudeiro et al. (2019b) were the first within the facial animation synthesis community to use speech features extracted from a pretrained text-to-speech model (DeepSpeech Hannun et al. 2014). Subsequent works in the field have used features extracted from a variety of popular models such as Whisper (Radford et al., 2022), WavLM (Chen et al., 2022), HuBERT (Hsu et al. 2021), and wav2vec (Schneider et al., 2019) and its variants (Baevski et al., 2020; Barrault et al., 2023). These pretrained models capture higher-level linguistic and other semantic audio features through large-scale self-supervised training. Because they are trained specifically on audio data that is closely related to the audio in A-roll footage, they enable downstream models to represent speaker identity as well as utterance content and context, typically allowing smooth generalization to unencountered contexts. This is beneficial for low-data tasks with a narrow scope, such as talking head generation, it can be limiting for audio-to-video tasks. A-roll generation depends on a wide range of signals including environmental cues, like background room tone, to distinguish indoor from outdoor settings as well as subtle paralinguistic sounds such as breath or ambient noise.

# METHOD

Mirage uses a Diffusion Transformer (DiT) to generate video tokens optionally conditioned on audio, text, and reference images. We now describe the architecture of the DiT model and the training losses used to train Mirage.

### 3.1. VIDEO ENCODING

Mirage uses Mochi's 3D spatiotemporal Variational Autoencoder (VAE) [(Genmo Team, 2024)](#) to represent videos in latent space. The VAE interleaves causal convolutional blocks and attention blocks to downsample the input video by a factor of 6 in time and a factor of 8 in both spatial dimensions. We encode videos in 25 frame chunks which yields a total of 5 latents that maintain the temporal causality of the video. These latents are then further convolved using a 3D convolution and patched with a patch size of 2 ×2 to produce the final video tokens. This translates to a vector of size 12 ×20 ×160 ×90 for a 4-second video sequence with 100 frames at 720p resolution. We flatten-patched video latents to a vector of sequence length 72000.

### 3.2. AUDIO ENCODING

We encode 16 kHz single channel audio using wav2vec and bilinearly interpolate the latents to match the video frame rate of 25 frames per second. We keep the original embedding dimension of size 1024. We crop segments of these latents to align with the video scene boundaries and then pass them through a learnable projection layer before they are used as input to the Transformer.

### 3.3. TEXT ENCODING

Mirage uses the pretrained T5-XXL model (T5 = Text-to-Text Transfer Transformer, [Raffel et al. 2020](#)) to encode text descriptions of each video. We obtain text descriptions using vision language models (VLMs). VLM-generated video descriptions, capped at 256 tokens, provide crucial scene context and visual content details (see [Sec. 3.7.4](#) for details). The T5 encoder outputs a dense vector representation of shape 256 ×4096 (sequence length ×channel dimension).

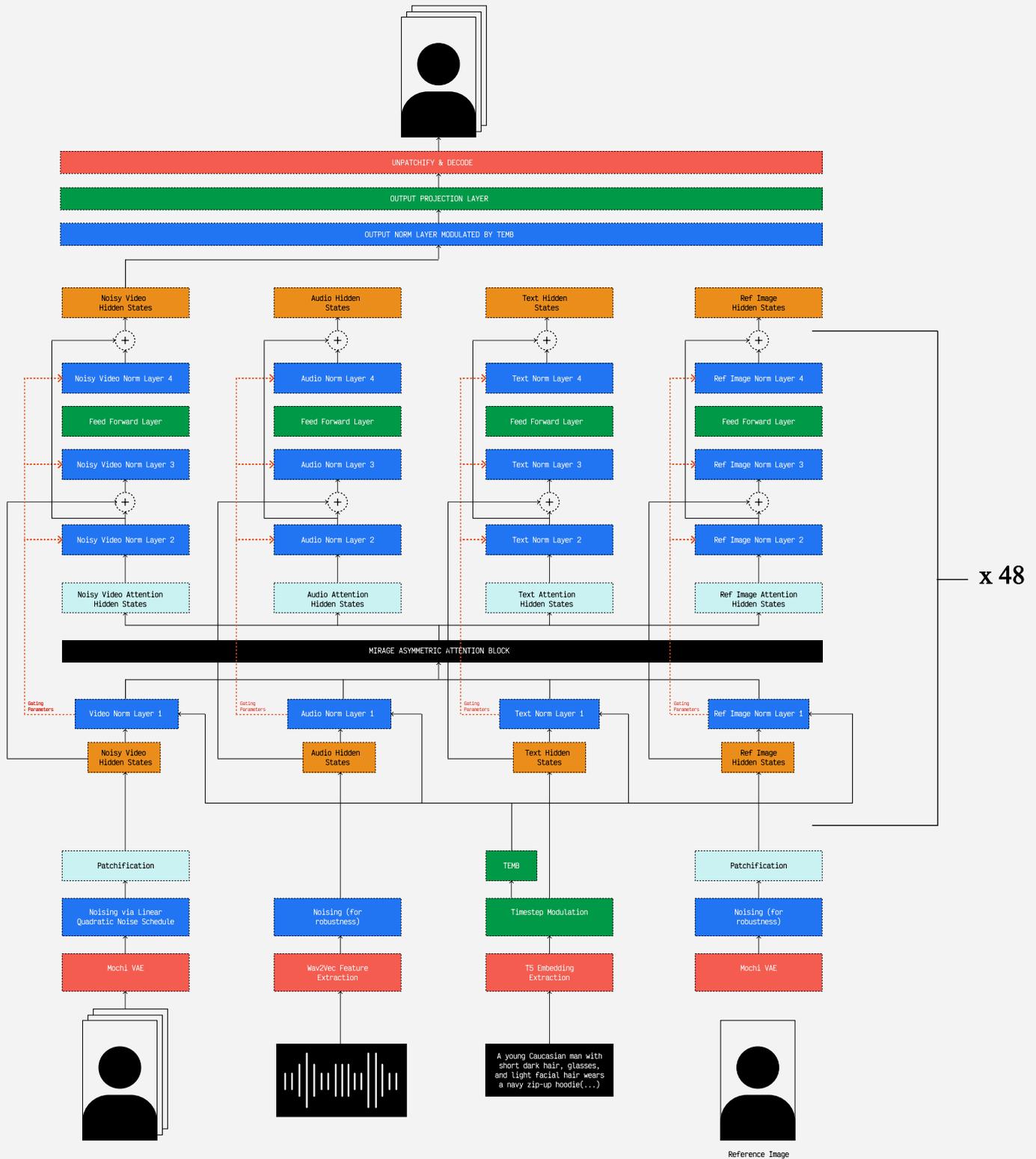

● Figure 2

The Mirage architecture. Video, audio, text, and reference images are first processed with modality-specific subnetworks to obtain latents for each modality. The text latents are modulated by the timestep to obtain a joint text-timestep embedding, as well as a separate timestep embedding. Each modality latent is then passed through 48 Mirage asymmetric self-attention blocks which consist of the attention operation itself as depicted in Fig. 3, as well as a number of normalization layers which each modulate their associated features. Gating parameters are used to modulate the information passed through residual connections.

## 3.4. ARCHITECTURE

Mirage's Transformer architecture is based on the Mochi asymmetric DiT (Genmo Team, 2024), an open-source text-to-video model. Mirage uses a 10-billion parameter DiT with 48 consecutive Transformer blocks that process audio, text, and video tokens in a joint self-attention operation. We illustrate the Mirage architecture in Fig. 2. Each Mirage Transformer block takes input audio, text, and video tokens and concatenates them along the sequence dimension. Mirage uses learnable Rotary Position Embeddings (RoPE, Heo et al. 2024) built on the original RoPE by (Su et al., 2024) and extended to 3-dimensions. Over the course of training, the network learns to produce a set of outputs that capture information across a range of spatiotemporal frequencies. The RoPE'd tokens are then used to calculate timestep-modulated bias, scale, and gating values using *adaLN-Zero* feedforward layers (Peebles and Xie, 2023). The scale and bias are then used to normalize each modality before they are processed by a multi-head attention mechanism that acts on the full sequence of [*text, audio, video*]tokens (See central block of Fig. 2). The output of self-attention module is then normalized and scaled using another set of feedforward layers and gated using the gating values that were calculated pre-attention. The central motivation for concatenating conditional modalities along the sequence dimension instead of using cross-attentions or sequences of specialized attention layers is its extensibility. Any *N* conditioning tokens can be concatenated along the sequence dimension allowing cross-modal interaction by design.

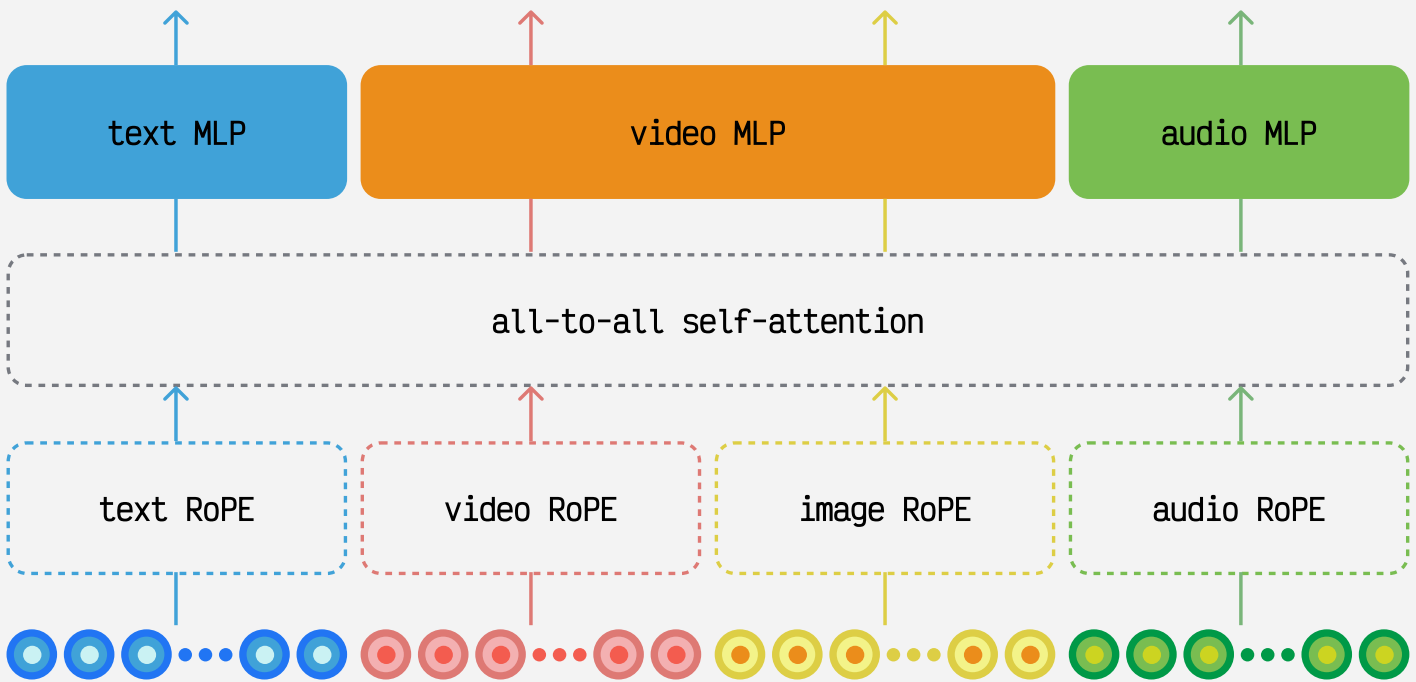

● Figure 3

Mirage asymmetric attention block. Mirage extends Mochi-style asymmetric self-attention to multiple modalities. Each modality is handled as a token stream of uniform dimensionality, processed with a modality-specific learned rotary position encoding (RoPE), and allowed to attend to all tokens from all modalities. Each modality is processed with a separate, modality-specific MLP. Note that all imagery (reference images and videos) are handled as a single modality.

### 3.5. TRAINING LOSS

We train Mirage using latent Flow Matching (Liu et al., 2023b), flowing from unit Gaussian noise $\mathbf{x}_0 \sim \mathcal{N}(0, I) \in \mathbb{R}^{T_l \times H_l \times W_l}$ to the latent data distribution $\mathbf{x}_1 \sim \mathcal{D} \in \mathbb{R}^{T_l \times H_l \times W_l}$, where $T_l \times H_l \times W_l$ are the spatio-temporal dimensions of the latent space. The velocity prediction at any timestep $t$ is conditioned on both audio and text $[\mathbf{a}, \mathcal{T}]$.

During training, we minimize the following loss

$$\mathcal{L}_{\text{RFM}}(\theta) = \mathbb{E}_{t, \mathbf{x}_1 \sim p_{\text{data}}, \mathbf{x}_0 \sim \mathcal{N}(0,I)} \left[ \|\mathbf{v}_\theta(\mathbf{x}_t, t, [\mathbf{a}, \mathcal{T}]) - (\mathbf{x}_1 - \mathbf{x}_0)\|_2^2 \right]; \text{ where } \quad \mathbf{x}_t = (1-t)\mathbf{x}_1 + t\mathbf{x}_0 \quad , \quad (1)$$

where $\mathbf{v}_\theta(...)$ is the velocity predicted by Mirage DiT with parameters $\theta$. We optimize $\mathcal{L}_{\text{RFM}}$ using the AdamW optimizer (Loshchilov and Hutter, 2019). We sample $t$ from a linear quadratic schedule during both training and inference.

### 3.6. REFERENCE IMAGE CONDITIONING

As described in Sec. 3.4, extending Mirage for other conditions is as simple as concatenating the new tokens along the sequence dimension. We do exactly that to condition on reference images. Our conditioning signal is now [*ref − image, text, audio*] with the full sequence being [*ref − image, text, audio, video*], the loss now is:

$$\mathcal{L}_{\text{RFM}}(\theta) = \mathbb{E}_{t, \mathbf{x}_1 \sim p_{\text{data}}, \mathbf{x}_0 \sim \mathcal{N}(0,I)} \left[ \|\mathbf{v}_\theta(\mathbf{x}_t, t, [\mathbf{ref}_l, \mathbf{a}, \mathcal{T}]) - (\mathbf{x}_1 - \mathbf{x}_0)\|_2^2 \right] \quad (2)$$

where, $\mathbf{ref}_l$ is the reference image latent extracted from the VAE as described in Sec. 3.1 by repeating the image 7 times in the temporal dimension. This is because the minimum input size for the VAE is 7 frames. During training, we sample the reference frame from a subsection of the video that is outside the range of the video currently sampled to compute the flow-matching loss. Since reference image latents are in the same domain as the video latents, we re-use the video latent specific parameters for the reference image latents as well, with the only exception of positional encoding, where we set the time value (time in the video, not flow-matching time) as the relative position of the reference frame to the video subsequence sampled for the flow-matching loss.

### 3.7. TRAINING DATA

We constructed a curated training dataset specifically for Mirage training. Each sample contains a normalized video stream in 720p portrait, 25fps H.264 paired with 16kHz mono-channel AAC encoded audio. To support multimodal conditioning and generation, each clip is used to compute a VAE embedding, a text embedding derived from the labeled prompt associated with the video, and an audio embedding capturing speech characteristics. Those representations enable the model to learn fine-grained correspondences between visual appearance, speech audio, and textual context.

### 3.7.1. DATA PROCESSING

#### ARCHITECTURE

To support model development, we built a custom large-scale data processing system responsible for transforming raw heterogeneous video samples into high-quality preprocessed training data, designed with an emphasis on adaptability, scalability, and performance.

Our data processing system follows an event-driven, asynchronous architecture comprised of loosely coupled components connected by a central real-time message exchange. Feature extractors form a bottom-up declarative workflow execution graph, where each feature extractor defines its upstream dependencies. We use a lightweight controller to orchestrate work across them and track execution states and resolve dependencies. This setup allows components to be deployed, configured, and scaled independently. We maintain high throughput and efficient resource utilization while enabling a responsive and incremental development process.

#### STORAGE

We use a two-tiered storage model to store and serve outputs. Features such as structured metadata and derived attributes are first written to online storage for downstream use and then later replicated to a data warehouse for training. We stored artifacts including large binary outputs referenced by features such as media assets and embeddings in persistent blob storage.

We implemented a unified task-locking and result storage mechanism that leverage atomicity guarantees from Google Cloud Storage (GCS, [Google Cloud 2025](#)). This vastly simplified the development process and improved resource efficiency by eliminating race conditions and duplicate work. The idempotent and fault-tolerant nature of our workloads lets us use pre-emptible or otherwise underutilized compute resources.

#### CHALLENGES

Working with video data at this volume introduces several infrastructure challenges:

- The heterogeneity of videos across format, content, quality, and other axes make it difficult to draw clear boundaries across data samples. We often defer to qualitative or heuristic-based approaches in evaluation.

- The large size of video files and derived features requires additional capacity, bandwidth, and latency considerations for both storage and network.

Video-based features for tasks like shot detection tend to be highly sensitive to changing heuristics, requiring us to remain flexible while preserving correctness guarantees. We intentionally avoid general-purpose execution frameworks in favor of maintaining fine-grained control over our data models and deployments. This allows us to maximize resource efficiency through workload-specific optimizations across every layer.

Operation at this scale reinforces the importance of defensive system design. Throughout the development lifecycle, we seek to avoid destructive assumptions, instead favoring preservation, redundancy, and incremental changes. We strongly adhere to idempotency and aggressive parallelism as foundational philosophies to ensure the level of throughput and resilience necessary for research to move quickly without compromising data integrity.

### 3.7.2. SINGLE-SPEAKER SCENES

Our training data consists of audio-visual examples spanning a wide range of content types, visual and audio styles, and camera perspectives. While many of these videos contain segments featuring a single individual speaking, they often also include multi-speaker interactions, background activity, unrelated activity, and B-roll footage without people. We use MediaPipe [(Lugaresi et al., 2019)](#) to isolate high-quality A-roll segments. We apply PySceneDetect [(Castellano, 2025)](#) to identify scene boundaries based on visual content changes. We then intersect the identified frame ranges with the scene boundaries to extract valid scenes. We discard scenes shorter than 2 seconds and divide longer scenes into non-overlapping 10-second chunks.

### 3.7.3. DATA FILTERING

To ensure the quality, consistency, and suitability of our training data, we apply a comprehensive set of filtering strategies across multiple modalities. These filters target various attributes such as video format, audio-visual alignment, motion characteristics, and visual artifacts including text overlays, screen splits, and screen overlays.

---

`Static frames`     To improve the efficiency and relevance of video data selection, we propose a filtering method based on the motion vector energy inherent in the H.264 video compression standard. This approach builds on the observation first introduced by Emu Video [(Girdhar et al., 2024)](#), which uses the mean energy of motion vectors for downstream metrics, evaluation, and filtering. Given that motion complexity correlates with the amount of predictive data encoded, we employ the average packet size of P-frames as an indirect but effective measure of motion vector energy. This information can be efficiently extracted using tools such as FFmpeg [(Tomar, 2006)](#). Our filtering criterion excludes videos if the mean packet size of P-frames falls below a predefined threshold. This strategy allows for the removal of low-motion or static content, thereby optimizing datasets for applications that rely on motion-rich video input.

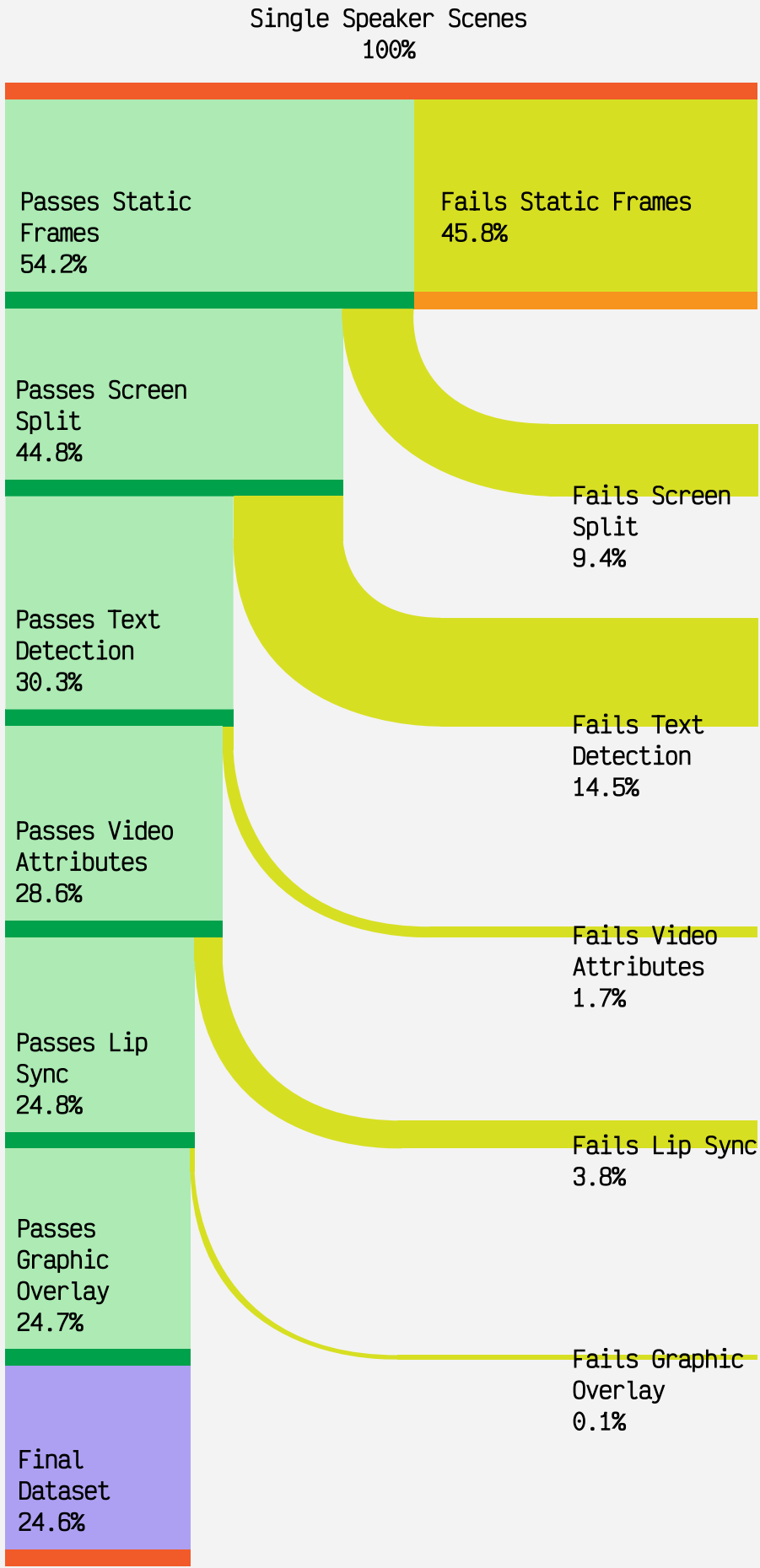

● Figure 4

Sankey diagram showing the progressive filtering of our dataset to produce the final set of training clips. Starting from single-speaker scenes, we apply a series of filters (e.g., static frame, screen split, text, video attributes, lip synchronization, and graphic overlay detection), resulting in a final dataset of high-quality video segments suitable for training.

| | |
|---|---|
| Screen-split detection | To detect screen-splitting artifacts in video data where creators divide their screen into multiple sections to show different content simultaneously, we apply the Canny edge detection algorithm (Canny, 1986), using the OpenCV library (Bradski, 2000). This method identifies prominent horizontal and vertical edges that often correspond to top-bottom screen divisions or overlaid elements such as text boxes and graphics. We then threshold over row and column statistics to determine if the video should be excluded. |
| Text detection | To filter out videos with excessive text overlays, we employ the CRAFT (Baek et al., 2019) text detection model through the EasyOCR (JaidedAI, 2024) implementation. For each video frame, we consider all detected text regions and calculate the total area covered by the bounding boxes. Videos containing text that covers more than 1% of any frame's area are excluded from the dataset, ensuring minimal graphic text overlay interference while preserving content with subtle or sparse textual elements. |
| Lip synchronization | We use the official SyncNet (Chung and Zisserman, 2016) framework to filter videos based on the quality of lip synchronization. SyncNet outputs two key metrics: (1) the offset value, which indicates the number of frames by which the audio and video are misaligned, and (2) the confidence score, which reflects the reliability of the synchronization estimation. We observe that videos that contain background noise or singing typically yield lower confidence scores despite having small offset values. By considering both the offset and the confidence score, we are able to include a wider range of videos under various acoustic conditions in our training set. For efficiency, if a video clip exceeds 10 seconds in length, we sample a 10-second segment for use in training uniformly at random. |
| Graphic overlay detection | To filter out videos containing overlaid graphics or brand logos from the collected footage, we develop an internal image-level graphic overlay detection model. We trained this model using a synthetically generated dataset, where various graphics and brand logos were overlaid onto image frames featuring talking individuals. To improve the robustness of the model, we applied a range of image-level augmentations during training, including Gaussian blur and pixel noise. We adopted the YOLOX |

architecture (Ge et al., 2021) and trained it from scratch on our synthetic dataset, following the original training recipe. This approach enables reliable identification and removal of videos with overlaid logos from the data pipeline.

### 3.7.4. VIDEO CAPTIONING

In our video captioning setup, each scene description is designed to include rich, identity-specific details to support realistic talking video generation. These captions go beyond static visual elements, capturing dynamic aspects such as hand gestures, facial expressions, and shot transitions.

● Figure 5

Example text description used for video captioning, illustrating detailed annotations of appearance, bearing, background, and shot composition.

| Attribute | Description |
| --- | --- |
| Appearance | A young Middle Eastern-North African woman with her curly dark brown hair tied back wears a gray knitted cardigan over a white top. |
| Bearing | She appears engaged and relaxed, looking directly at the camera and using dynamic hand gestures while speaking. |
| Background | The background features a bookshelf with a decorative plant and a vase, creating a tasteful, modern, professional atmosphere. |
| Shot | The shot is a medium, stationary shot framed at eye-level, illuminated by soft indoor lighting that provides clear visibility without harsh shadows. |

We annotate each video frame in the dataset using a standardized descriptive format that emphasizes clarity, precision, and consistency across scenes. We structure descriptions into four core components: appearance, bearing, background, and shot, each capturing distinct visual aspects of the subject and environment. The format begins with a confident, physical profile, followed by a detailed account of facial expression and gesture, then a thorough scene description, and finally a technical breakdown of shot composition and lighting. This method enables high-fidelity textual representation of visual content, supporting tasks in video understanding
and multimodal learning.

To generate these structured descriptions, we employ a multi-stage pipeline that first decomposes the visual content into static and temporal components. For the static aspect, we use the GPT-4o (OpenAI et al., 2024) and Qwen 2.5-VL (Bai et al., 2025) VLMs to analyze keyframes sampled at the start, one-third, two-thirds, and end positions of the video. We prompt the VLM to extract fine-grained, identity-specific attributes including facial appearance, hairstyle, clothing, background objects, and shot framing. These outputs are formatted in a JSON schema that captures both static descriptions and temporal changes in appearance, lighting, and scene composition.

We apply a second VLM pass to process a dense grid of keyframes sampled in every 1 second and extract dynamic scene patterns such as body movement, facial expression shifts, and camera transitions. The temporal-structured prompt follows a structured, hierarchical format that captures high-level action rhythms, fine-grained gestural changes, and timestamp-aligned motion sequences.

Finally, we feed both structured static and temporal prompts into the GPT-4o LLM (OpenAI et al., 2024), which synthesizes them into a single four-sentence paragraph following the standardized format shown in Fig. 5. This approach enables scalable, automated caption generation that is both visually grounded and semantically rich.

# MODEL TRAINING & INFERENCE

## 4.1. MODEL SCALING AND TRAINING OPTIMIZATION

### 4.1.1. PARALLELISM

We employ Fully Sharded Data Parallelism (FSDP) as our parameter sharding strategy. We achieved 100% computation-communication overlap in our hardware configuration.

The Mirage model input has dimensions $[b, s, d]$, where $b$ represents batch size, $s$ denotes sequence length, and $d$ indicates hidden dimension size. For long-context scenarios when sequence length $s$ extends to hundreds of thousands of tokens, attention layers become the primary computational bottleneck. To address this challenge, we implement context parallelism (CP), which partitions inputs along the $s$ dimension across multiple devices.

Context parallelism can be implemented through various approaches, including Ring Attention (Liu et al., 2023a), Ulysses (Jacobs et al., 2023), or hybrid methods such as USP (Fang and Zhao, 2024). We adopt Ulysses to enable efficient processing of long sequences distributed across multiple GPUs, overcoming the memory and computational constraints of individual devices. This approach shards the sequence length $s$ across the CP group and performs two all-to-all communication operations during attention computation. The first all-to-all operation reconstructs the original token sequence and reshards it along the number of attention heads $h$ dimension, while the second all-to-all operation gathers the head dimension hand reshards the sequence along the sequence dimension $s$. Together, these operations reduce both computational requirements and activation memory by a factor equal to the CP group size.

### 4.1.2. THROUGHPUT OPTIMIZATION

We leverage bfloat16 mixed-precision training to balance computational efficiency with numerical stability (Burgess et al., 2019). We observe that incorporating *torch.compile* yields approximately 35% throughput improvement over the baseline configuration (Paszke et al., 2019). We integrate FlashAttention-3 (Shah et al., 2024) for an additional 30% performance gain.

We apply selective gradient checkpointing to reduce memory usage during training. This technique stores only subset of intermediate activations and recomputes the others when needed, striking an effective balance between memory savings and computational cost. Our implementation reduces memory utilization while maintaining efficient training speeds.

### 4.1.3. FAULT TOLERANCE & MONITORING

As the number of GPUs in a process group grows, the probability of node failure rises correspondingly. To maximize effective training time for the Mirage model, we have a comprehensive fault tolerance strategy. We establish a structured on-call rotation to continuously monitor the training state, supported by an automated notification system that immediately alerts the designated on-call engineer of training failures. Additionally, we have advanced health-checking mechanisms that proactively identify potential GPU issues through performance metrics and error signature analysis. When node failures occur, the system automatically replaces the compromised hardware with a healthy node from the resource pool, minimizing downtime and ensuring training continuity with minimal manual intervention.

## 4.2. INFERENCE

Compared to training, model inference involves multiple sampling steps, we use 64 in our base case. To reduce the time required for each individual step we employ distributed inference and quantization. Furthermore, because attention layer outputs within the same DiT block are similar between the sampling steps in the latter stages of sampling, there is a substantial similarity between the conditional and unconditional outputs of Mirage. We use this knowledge to decrease the overall computational load.

### 4.2.1. PARALLELISM

To optimize memory efficiency and processing speed, we employ the same distributed architecture setup used during training. We use Fully Sharded Data Parallelism (FSDP) for model weight sharding across multiple GPUs, reducing per-device memory requirements. For long context generation, we implement context parallelism (CP) using Ulysses (Jacobs et al., 2023) to address the attention computation bottleneck. Our benchmarks demonstrate near-linear scaling efficiency with respect to the CP group size, effectively reducing inference time proportionally to the number of GPUs in the CP group.

### 4.2.2. QUANTIZATION

Our experiments reveal that FP8 precision for matrix multiplication (matmul) operations introduces minimal visual quality degradation while providing performance benefits. We apply FP8 quantization to all matmul operations within the DiT architecture, leading to a 5% reduction in inference time compared to the BF16 baseline.

### 4.2.3. INFERENCE TIME CACHING

During the analysis of the model we found the following:

- CFG similarity: In the later stages of sampling there is a substantial similarity between conditional and unconditional outputs of Mirage.

- Attention layer outputs within the same block are similar between sampling steps.

In line with these observations, we observed that a time-caching configuration similar to the FasterCache framework (Lv et al., 2025) could be used to speed up inference with no loss in perceptual quality. This caching results in 40% reduction in inference time.

### 4.2.4. IMPROVING SAMPLING QUALITY

We apply dropout on all conditioning modalities during training to allow us to use classifier free guidance (CFG) when sampling the model (Ho and Salimans, 2022). We find that techniques such as Spatiotemporal Skip Guidance (STG) (Hyung et al., 2024), negative text prompting, and cosine annealing the CFG scale generate improved results over vanilla CFG.

# RESULTS

We evaluate Mirage generated output videos on several key criteria:

#### PROMPT ADHERENCE

```
Evaluates how accurately the video reflects the given prompt, including
the appearance, background elements, and props. This also encompasses
the style of the video, such as camera movements, shot types, and
other cinematic details that match the intended look and feel.
```

### SUBJECT FACIAL DETAILS

Assesses the precision of facial movements, including the synchronization of the eyes, mouth, and audio. It also evaluates the generated expressiveness, ensuring that expressions correspond to the emotions and content of the speech.

### SUBJECT BODY MOTION

Focuses on the realism and relevance of the subject's body movements, including hand gestures, posture, and overall physical interaction. This category also looks at how well the gesticulations convey semantic meaning in relation to the speech, ensuring that movements align with both verbal and non-verbal communication.

In our work, we prioritize human-centric evaluation through visual inspections of generated outputs to holistically assess model performance. For tasks like video generation, success hinges on qualitative and subjective factors such as storytelling fluidity, emotional impact, and perceptual realism. We encourage readers to explore our exhaustive video gallery available here, which showcases Mirage's outputs in a wide range of scenarios.

## 5.1. PLOSIVE AND VISEME DYNAMICS

Here we showcase the articulation accuracy for plosive sounds, emphasizing temporal alignment with audio waveforms and physical realism in lip tension. There are six plosive consonants commonly used in English: p, t, k, b, d, and g. The sounds /p/ and /b/ are produced by bringing both lips together (bilabial), while /t/ and /d/ involve the tongue making contact with the ridge just behind the upper front teeth (alveolar). For /k/ and /g/, the back of the tongue presses against the area between the hard and soft palate (velar).

Mirage demonstrates exceptional precision in articulating plosive sounds, consistently producing highly accurate lip movements across variations in language, tone, volume, and visual appearance. Fig. 6 showcases a series of still frames that illustrate this capability. In the examples below, the audio is entirely generated. Each frame in the figure corresponds to the starting frame of a different plosive. For even more challenging examples that demonstrate plosive behaviour in the form of tongue twisters, please see Fig. 9.

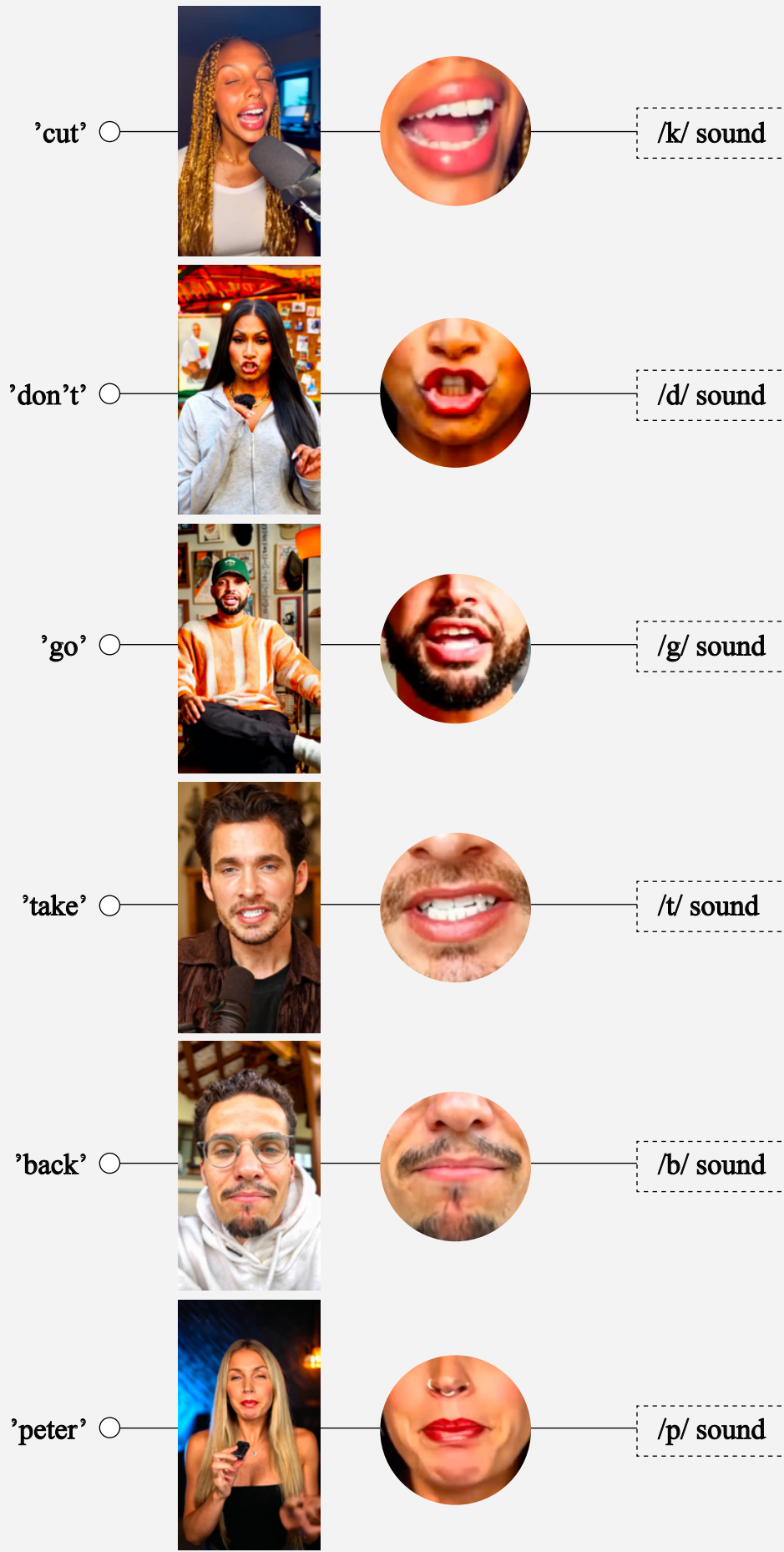

● Figure 6

Example Mirage outputs demonstrating typical behavior when conditioned on audio containing an audible plosive. Each frame in the figure corresponds to the starting frame of a different plosive

## 5.2. EYE BLINKING AND GAZE BEHAVIOR

Here we evaluate the model's capabilities at naturalistic eye movements, including blinks, saccades, and focal shifts tied to speech context (e.g., upward gaze during recall, sustained eye contact during assertions).

While challenging to measure quantitatively, empirical observations indicate that Mirage generates remarkably natural and context-sensitive eye behaviors. Blinks appear at realistic intervals, saccades are frequent yet nuanced, and gaze direction often reflects the underlying speech context. Despite not being trained to model blinking and eye motion explicitly, the model is able to synthesize these behaviors in a way that aligns with human expectations. This suggests that the learned audio-visual representations capture subtle correlations between speech patterns and eye dynamics, resulting in lifelike visual performances that enhance the overall realism of the generated videos.

Fig. 7 showcases examples of these dynamics, with each row representing a different video generated from distinct prompts and random seeds. Each frame depicts instances of blinking throughout the video.

● Figure 7  Mirage outputs feature natural blinking and eye movement.

A.
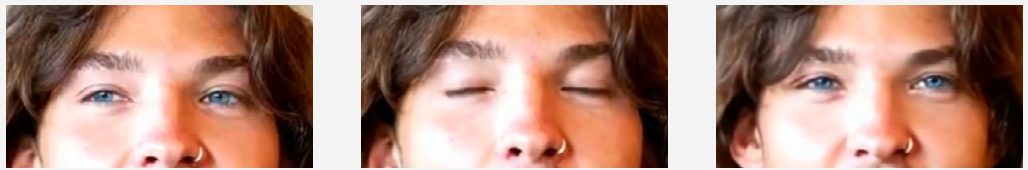

B.
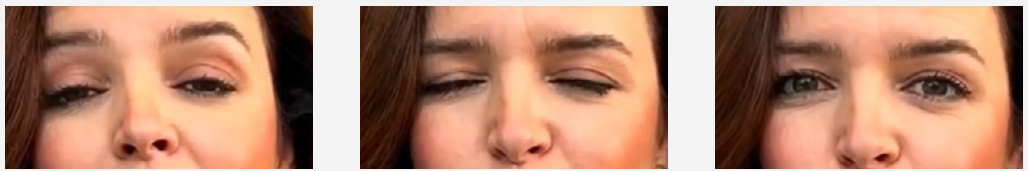

C.
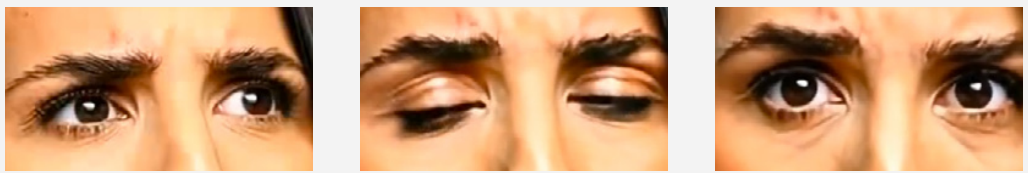

## 5.3. EMOTIONAL NUANCE AND AFFECT MATCHING

Here we demonstrate how facial expressions in videos generated by Mirage dynamically adapt to emotional cues present both in speech as well as described through the text prompt (e.g., eyebrow raises for surprise, lip curvature shifts for sarcasm), ensuring micro-expression timing aligns with vocal prosody.

Fig. 8 presents a series of still frames that demonstrate Mirage's ability to generate emotional videos from synthetic speech input using prompt control. Each row corresponds to a different emotion, with all images in a row corresponding to the same video, varying only by the emotion described in the prompt.

● Figure 8  Example Mirage outputs demonstrating typical behavior when prompted for different emotional states.

Mirage behavior when expressing happiness

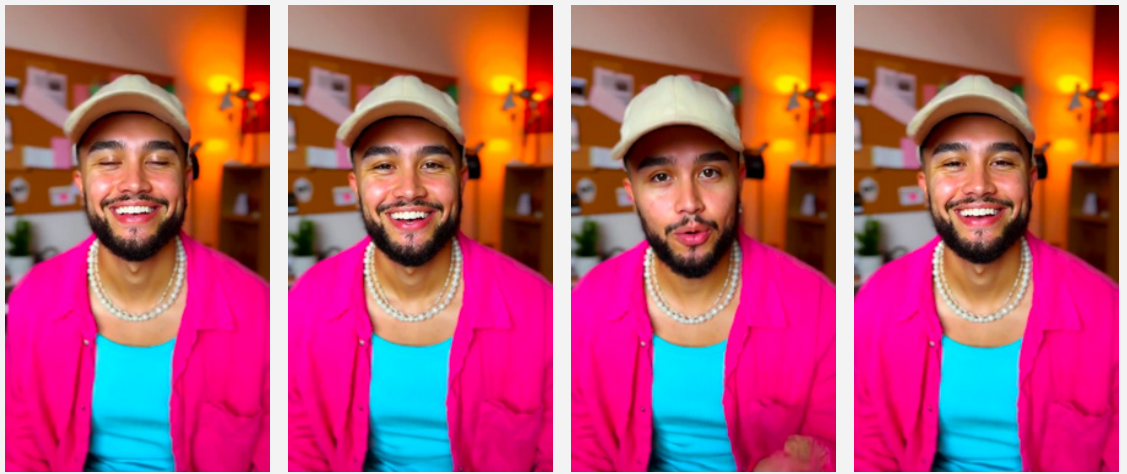

Mirage behavior when expressing sadness

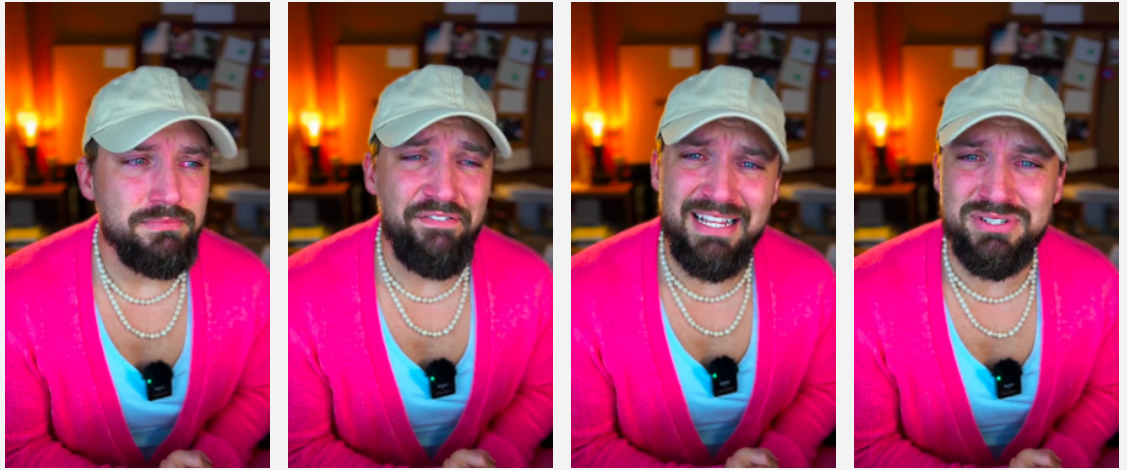

Mirage behavior when expressing anger

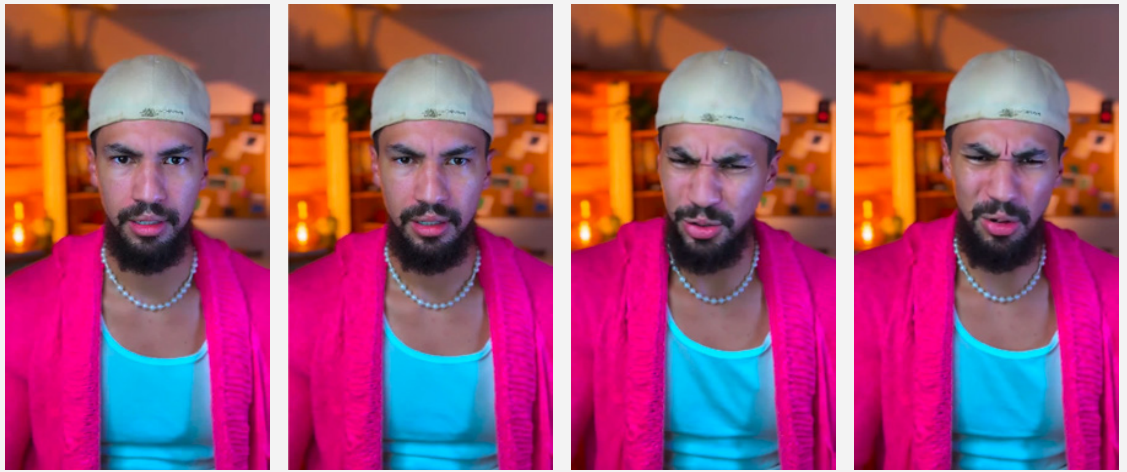

Mirage behavior when expressing neutral/flat emotion

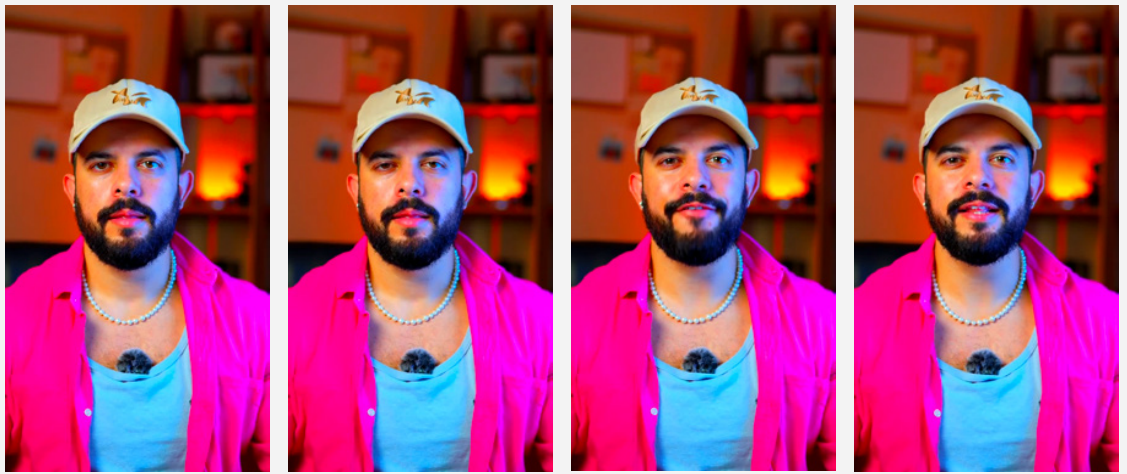

Mirage behavior when expressing excitement

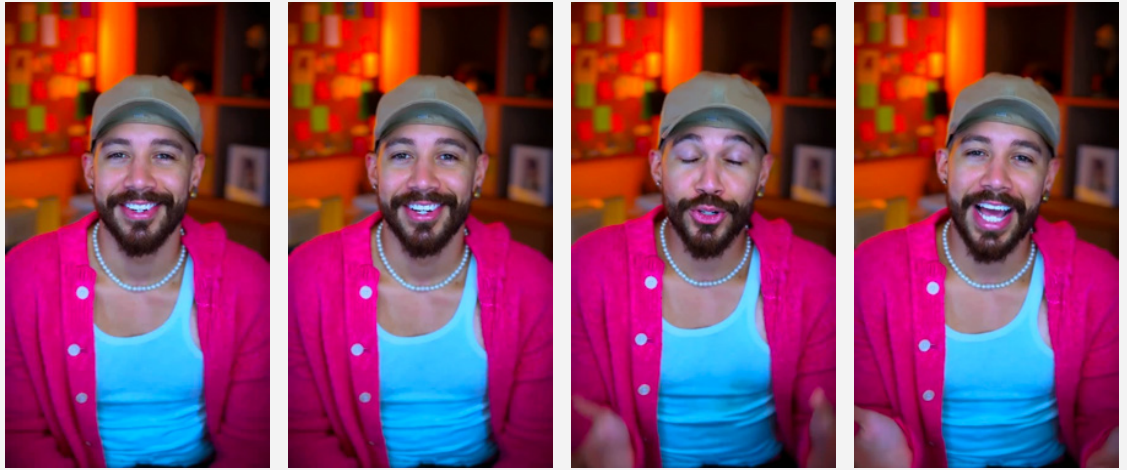

Mirage behavior when expressing shock

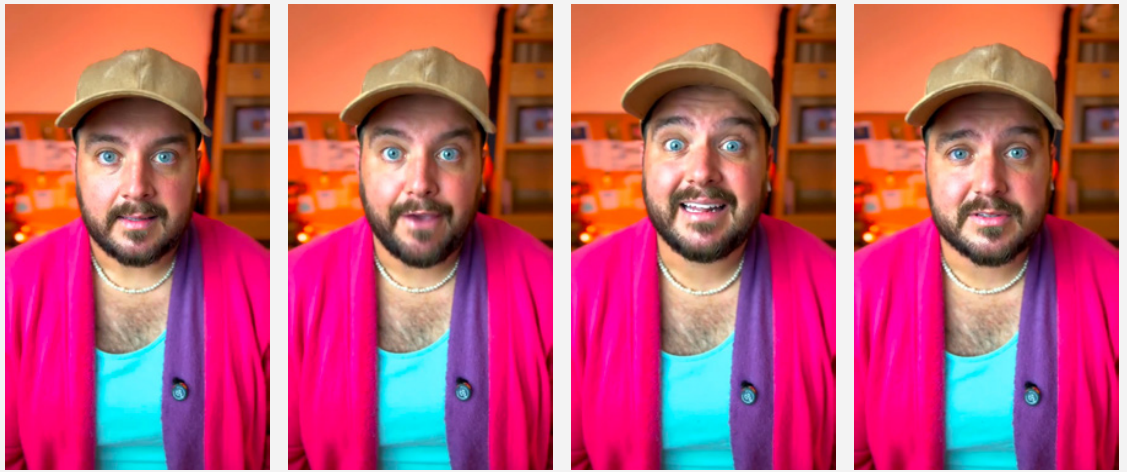

Mirage behavior when expressing fear

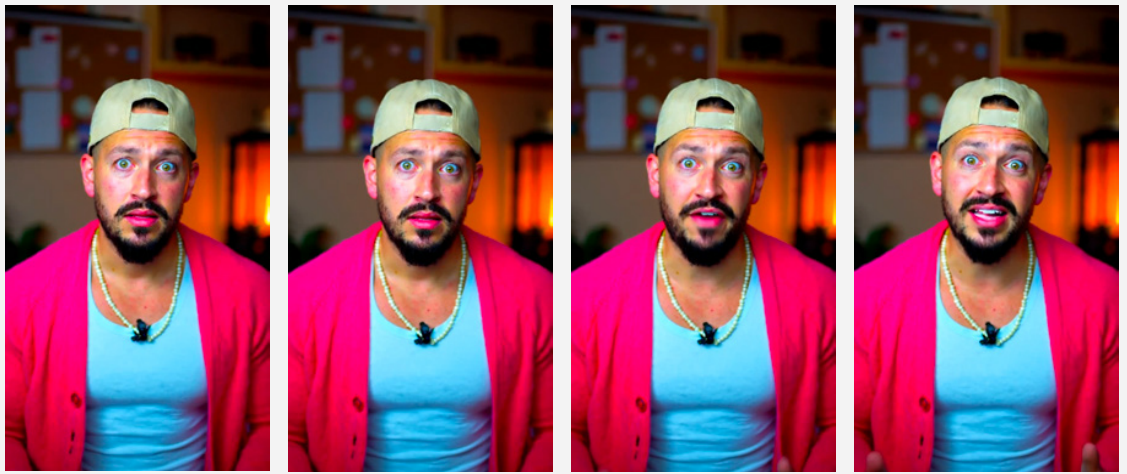

## 5.4. CO-ARTICULATION AND SPEECH BLENDING

Here we assess Mirage 's ability to generate smooth and natural transitions between phonemes in the context of tongue twisters, which present a unique challenge due to their rapid articulation and complex phonetic sequences.

Mirage shows strong performance in modeling coarticulation, blending visemes in a way that mirrors the continuous and anticipatory movements seen in natural human speech. Instead of producing rigid or segmented mouth shapes, the model transitions fluidly between articulatory targets, preserving realism even under the demanding timing of tongue twisters.

Fig. 9 highlights this with close-up crops showing the articulation of particularly difficult phoneme combinations in a range of different settings. Many of these sequences include fast-paced alliteration and minimal pairs that place high demands on timing and mouth coordination.

Despite these challenges, Mirage maintains clear and consistent mouth motion, correctly aligning lip closures, openings, and transitions with the accompanying audio. This indicates that the model captures subtle phonetic and rhythmic cues, allowing it to generalize effectively to fast and phonetically complex speech. Notably, the model also produces accurate tongue movements learned directly from data, without relying on explicit supervision. This is particularly striking given prior work that explicitly models tongue motion from speech (Medina et al., 2022).

● Figure 9    Mirage samples exhibit strong co-articulation across diverse speech patterns, such as tongue twisters.

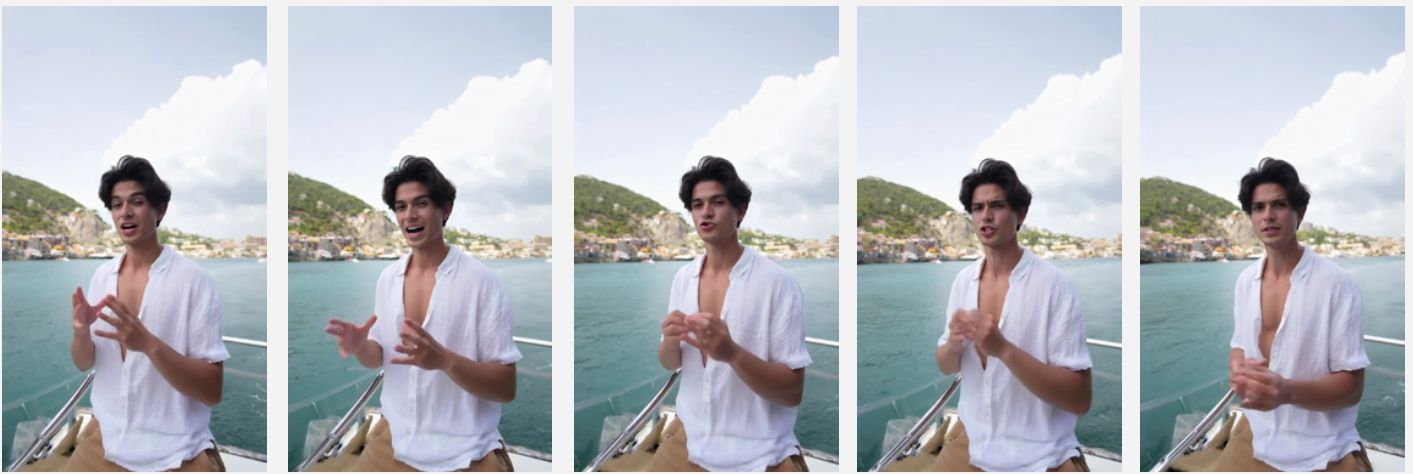

## 5.5. PARALINGUISTIC GENERATION

Mirage is capable of modeling non-lexical verbalizations such as laughter, sneezing, coughing, and sniffling. Fig. 10 depicts a handful of such examples.

Mirage behavior when audio features coughing

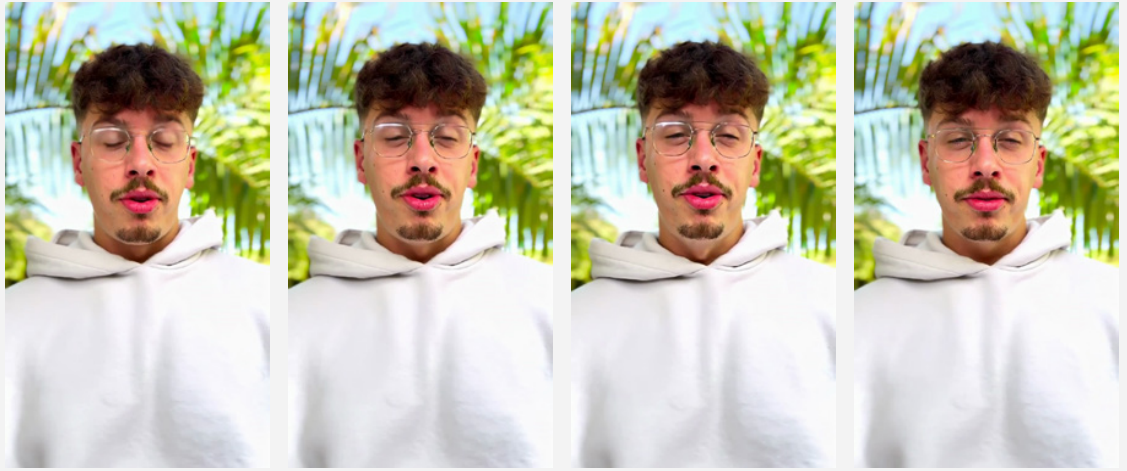

Mirage behavior when audio features laughing

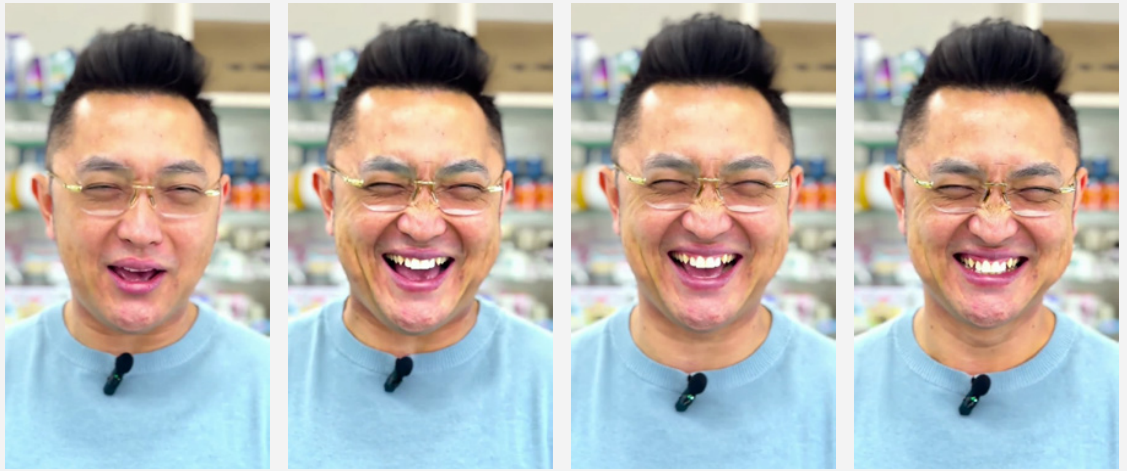

Mirage behavior when audio features sneezing, and sniffling

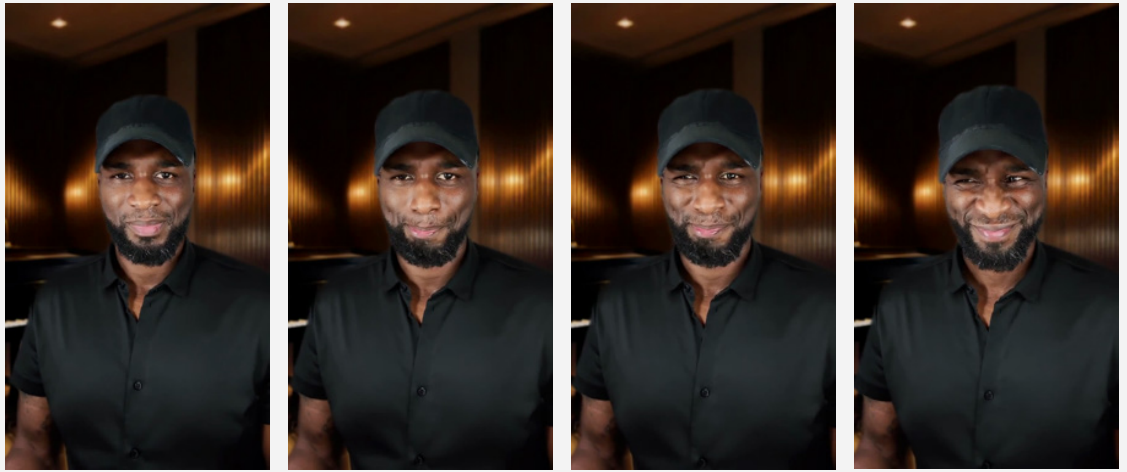

● Figure 10   Generated responses to a range of paralinguisitc sounds

## 5.6. AUDIO-ONLY CONDITIONING

When text prompts are omitted, Mirage correlates speaker attributes and environmental context directly from audio signals, leveraging latent correlations between acoustic properties and visual semantics. The model demonstrates an emergent ability of reconstructing ambient scene details from audio alone. Note that the visual fidelity of audio-only generations is diminished by comparison to text-audio generated videos. We now highlight interesting details that the model has been able to learn from the audio. For instance:

● Figure 11  Example Mirage outputs demonstrating typical behavior in response to male- and female-presenting voices.

Mirage output conditioned on male-presenting voice without text prompt
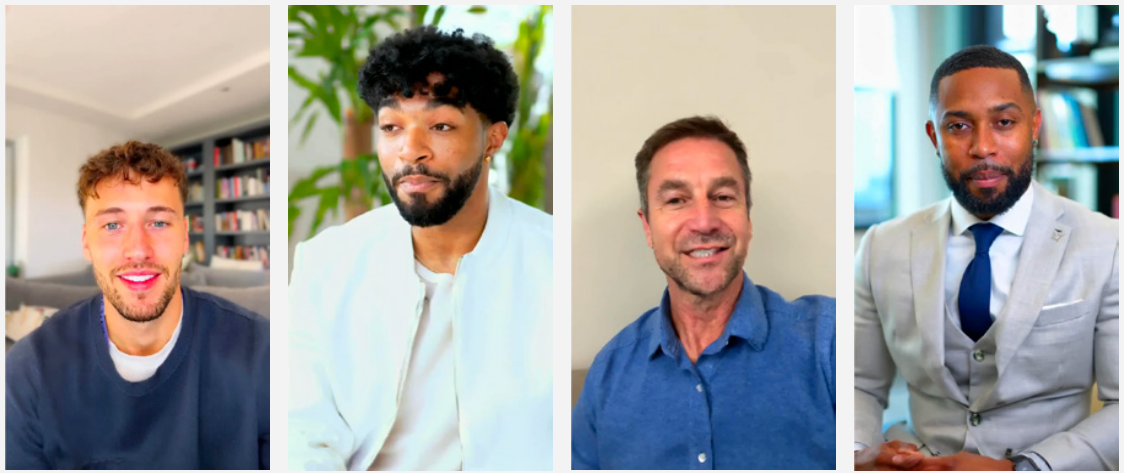

Mirage output conditioned on female-presenting voice without text prompt
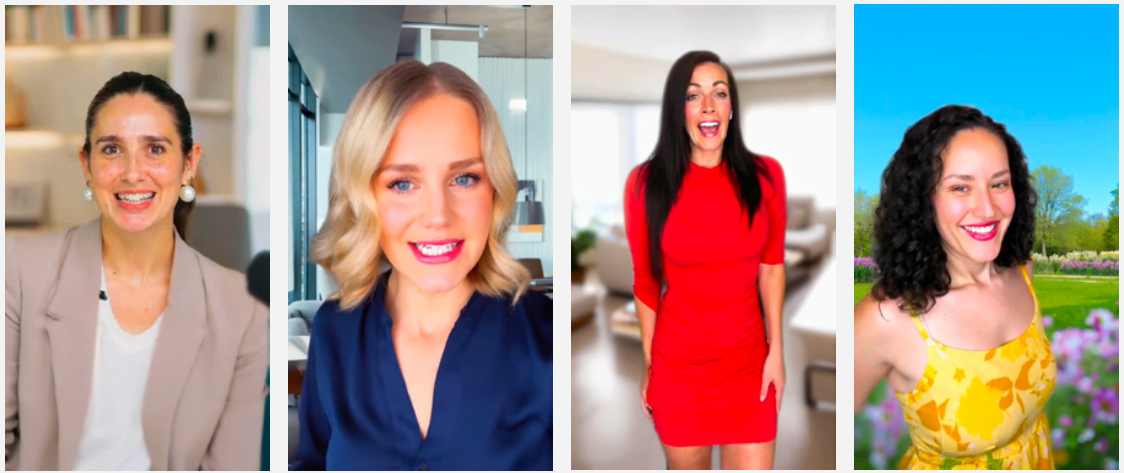

The model most likely correlates correct environmental context by analyzing reverberation profiles and background noise. Studio-grade audio (clean, dampened reverb) generates neutral, well-lit indoor environments, while "in-the-wild" audio (e.g., café chatter, wind interference, or traffic sounds) produces outdoor scenes. For a number of such examples, please see Fig. 12.

● **Figure 12**  Example Mirage outputs demonstrating typical behavior in response to audio with acoustic properties characteristic of some environment.

**Mirage output conditioned on speech recorded indoors without text prompt**

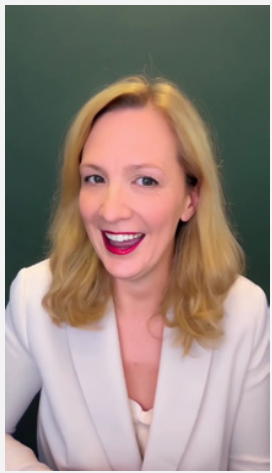 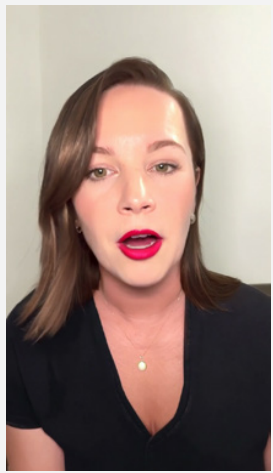 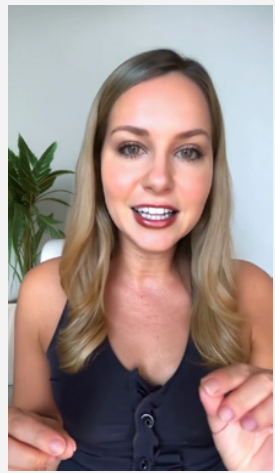 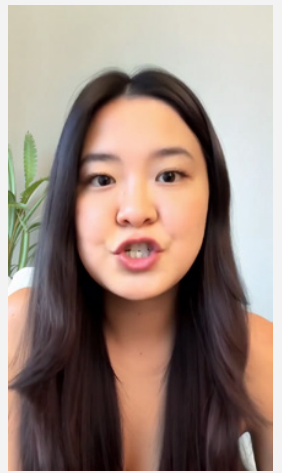

**Mirage output conditioned on speech recorded outdoors without text prompt**

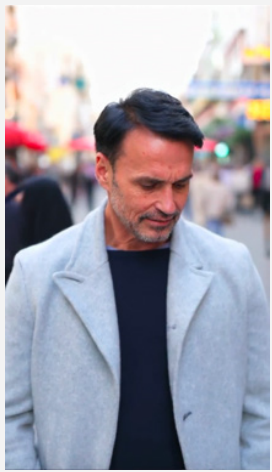 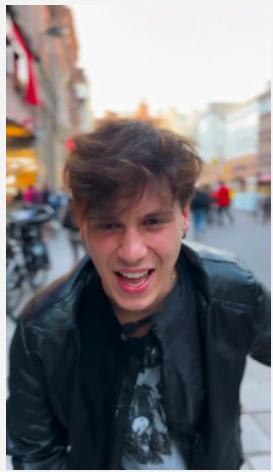 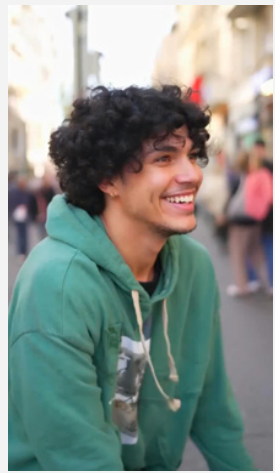 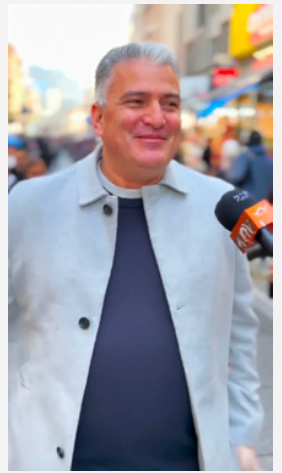

Mirage generates visually coherent appearances conditioned on vocal input, effectively avoiding the uncanny valley effects often observed in traditional talking-head synthesis, which can result from mismatches between a speaker's visual appearance and vocal characteristics. Fig. 13 depicts stills from several such videos. Note that all the videos were generated using the audio-only setting of the model, as such sample quality is degraded compared to join text-audio generation.

● Figure 13   Example Mirage outputs demonstrating typical behavior in response to acoustic signals from a wide range of people.

Mirage output conditioned on a wide array of voices

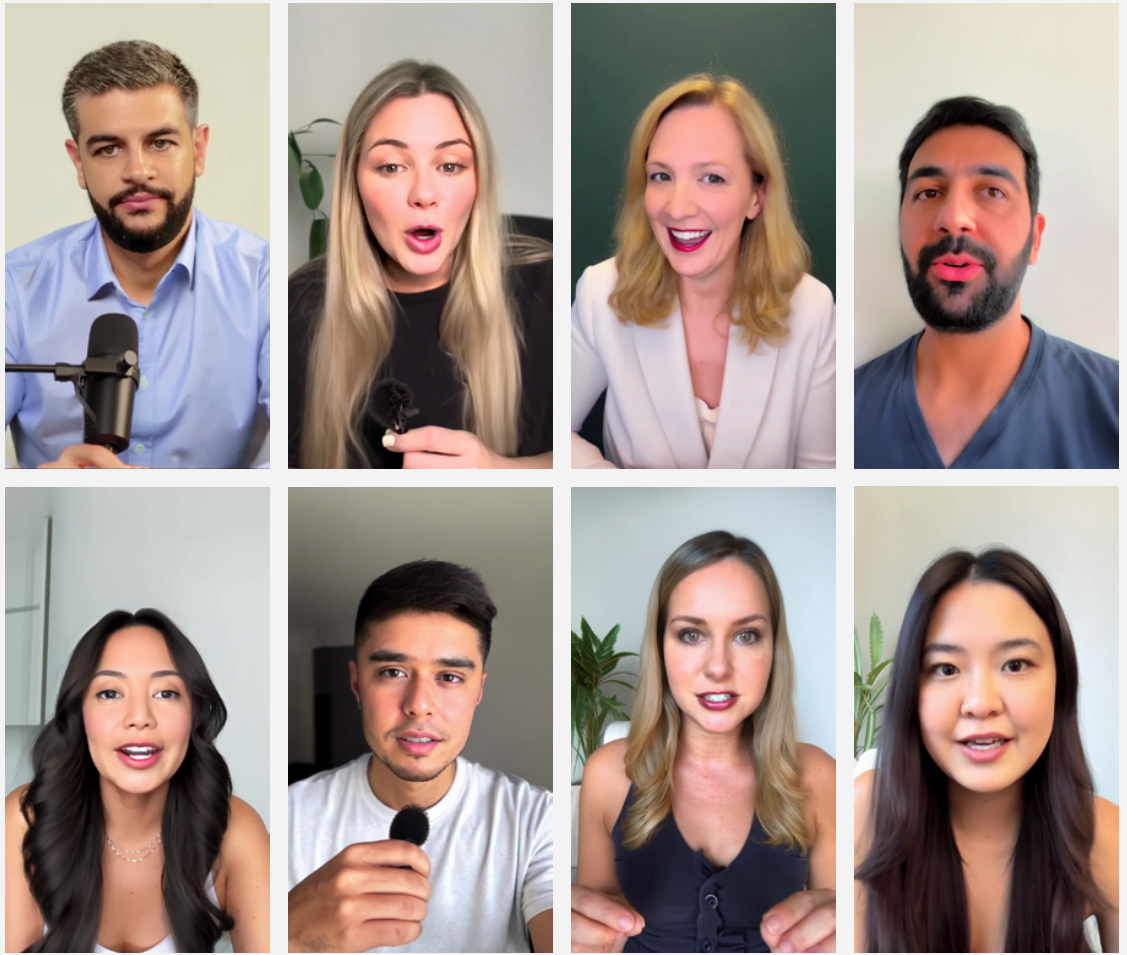

## 5.7. MISMATCHED TEXT AND AUDIO RESULTS

So far, we've focused on how Mirage can correlate scene-specific properties from audio alone. Sometimes cues derived from the audio may contradict the content of the text prompt. Here, we report Mirage 's behavior under such conflicting conditions by pairing text prompts with semantically mismatched audio.

Overall, the model is robust to mismatched inputs. Minor discrepancies such as audio recorded outdoors paired with a text prompt describing an indoor scene are handled gracefully. More pronounced mismatches, like a prompt describing a male-presenting subject accompanied by a female-presenting voice, often lead to outputs that visually lean toward the vocal characteristics. In this case, for instance, the generated subject may appear with longer hair, softer facial features, and other traditionally feminine traits. In Fig. 14, we showcase a number of examples where the voice and prompt description mismatch and the model's attempts at interpolating between the two.

We consistently observe that the most perceptually realistic results appear when the text and audio are semantically and affectively aligned. This pattern holds across a wide range of scenarios. When audio and text are in harmony, Mirage is able to produce highly coherent, contextually rich visual performances. When they diverge, the model still produces plausible results, but the output tends to be more ambiguous or stylized, suggesting a form of averaging or interpolation negotiation between the results suggested by each modality in isolation.

● Figure 14  We highlight Mirage results when the prompt and audio condition are mismatched. Each video is created using a unique random seed with mismatching text and audio.

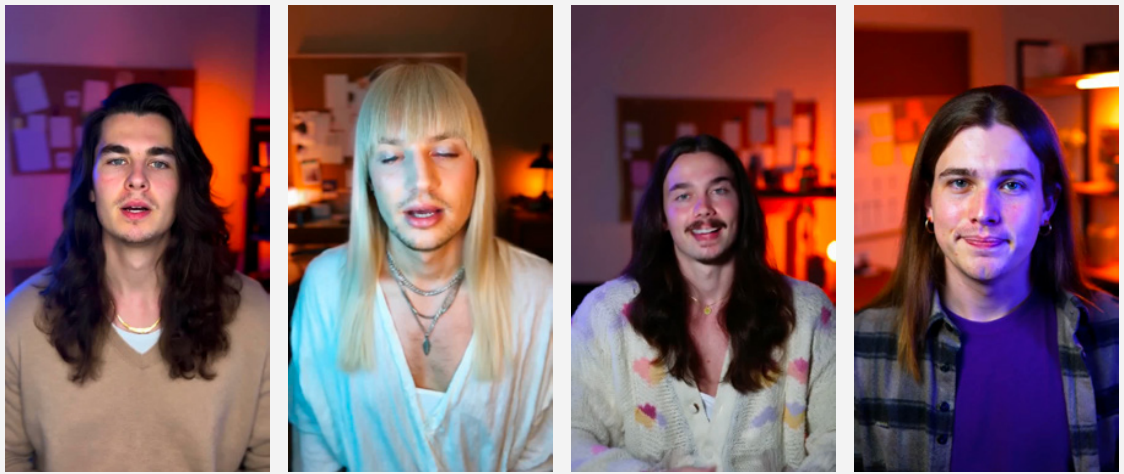

Mirage output when conditioned using a male-presenting voice and female-presenting text prompt

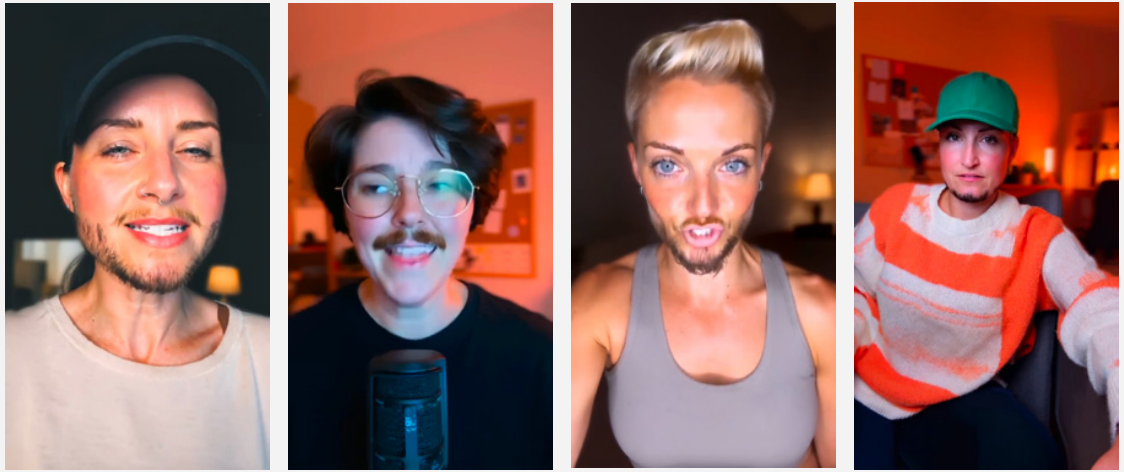

Mirage output when conditioned using a female-presenting voice and male-presenting text prompt

## 5.8. GESTURE-SEMANTIC ALIGNMENT

We now highlight instances where gestures appear meaningfully aligned with the spoken content. For example, nodding in response to agreement, or outward palm gestures during explanations. While this type of behavior is extremely difficult to quantify, we present empirical examples where such alignment clearly emerges.

The underlying mechanism remains unclear: it's difficult to determine whether the model is learning high-level semantic patterns from the wav2vec features, associating specific vocalizations with particular gestures, or simply producing coincidental but convincing outputs. Regardless of the cause, the resulting gestures consistently appear natural, expressive, and well-timed, contributing to the realism of the generated videos.

Fig. 15 illustrates cases of these semantically synchronized gestures, including head nods coinciding with verbal affirmations, subtle head shakes during expressions of uncertainty, and purposeful hand movements that punctuate storytelling or explanations. The timing and fluidity of these motions enhance the perception of intentional communication, such as a speaker raising their palms outward when describing abstract concepts or tilting their head thoughtfully during pauses. While the system's ability to generate context-aware gestures remains speculative, the outputs frequently mirror human conversational conventions. Though the model's "understanding" of semantics is unproven, the empirically observed alignment suggests an emergent capability to map vocal patterns to nonverbal cues, offering compelling implications for applications in virtual avatars and human-computer interaction. Further analysis is needed to disentangle learned associations from stochastic pattern matching, but the realism achieved here marks a notable advance in gesture synthesis.

● Figure 15 Example Mirage outputs demonstrating typical behavior in response to implicit signals of agreement and disagreement.

Head nodding in response to agreement. Prompts contained no explicit motion references.

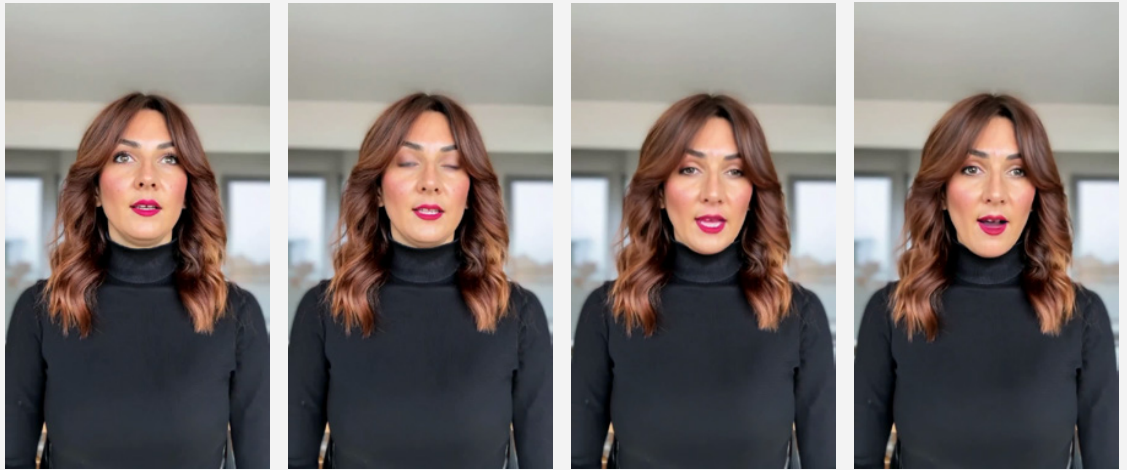

Head shaking indicating disagreement. No mention of body motion in prompts.

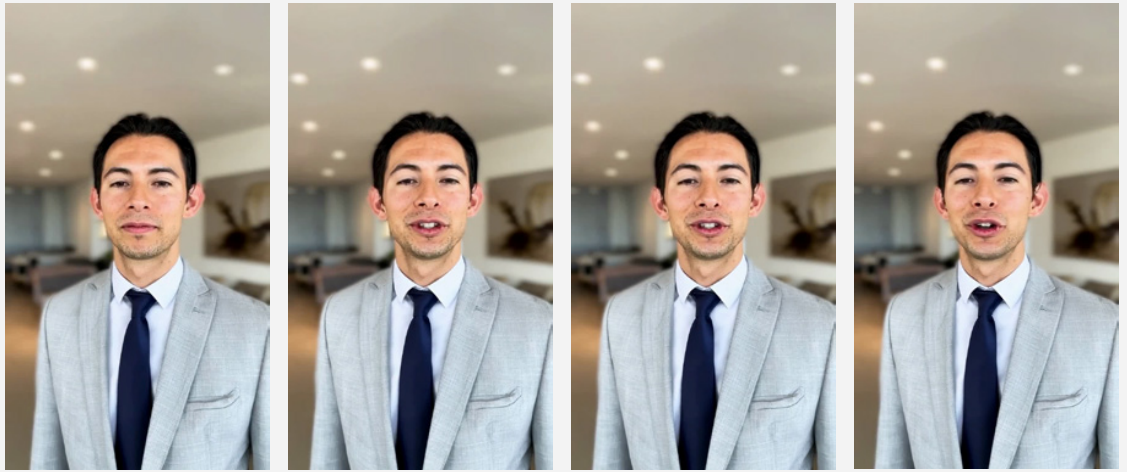

Purposeful hand movements in response to explanations. No mention of body motion in prompts.

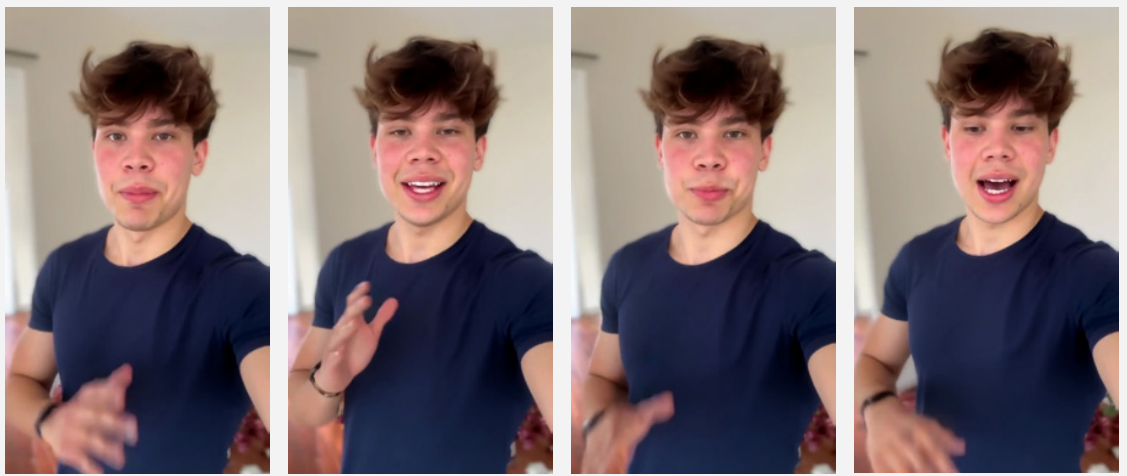

## 5.9. SUBJECT APPEARANCE ACCURACY

We next validate Mirage output alignment with physical descriptors and implicit traits (e.g., "a tired scientist") and demonstrate robustness to ambiguous prompts (e.g., "a charismatic leader").

Fig. 16 presents a series of subject appearance prompts with each row featuring a slightly different variation, alongside corresponding frames from the generated video. This illustrates how Mirage effectively integrates new descriptors into the appearance prompt while maintaining consistency and visual coherence across outputs. The results highlight the model's flexibility in adapting nuanced changes without compromising the subject's core identity.

● **Figure 16**  Select video frames and corresponding appearance prompts produced by Mirage.

```
A young woman with a long, dark, messy braid and silver hoop
earrings, her lips slightly parted, exuding a relaxed and mini-
mal aesthetic. She engages with the camera in a calm manner, her
expression thoughtful as she occasionally shifts her gaze, in-
viting viewers into her world. The background is softly blurred,
featuring a minimalist indoor space with neutral-colored walls, a
simple wooden table, and a few carefully placed decorative items
that enhance the serene atmosphere. The shot is a medium, sta-
tionary shot at eye level, under bright, even lighting that cre-
ates a clear and engaging visual of the subject.
```
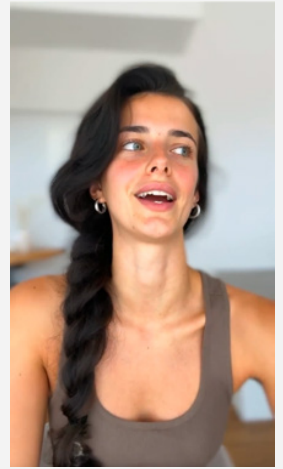

```
A young woman with a long, dark, messy braid cascading over her
shoulder, adorned with large hoop earrings, and sporting long,
manicured nails that add a touch of elegance to her look. She
engages with the camera, her lips slightly parted as if mid-sen-
tence, conveying a sense of openness and approachability while
maintaining a relaxed demeanor. The background is softly blurred,
featuring a minimalistic indoor setting with neutral-colored
walls, a few carefully placed decorative items, and soft tex-
tures that enhance the serene atmosphere. The shot is a medium,
stationary frame at eye level, under bright, even lighting that
creates a clear and engaging visual of the subject, highlighting
her features and the subtle details of the environment.
```
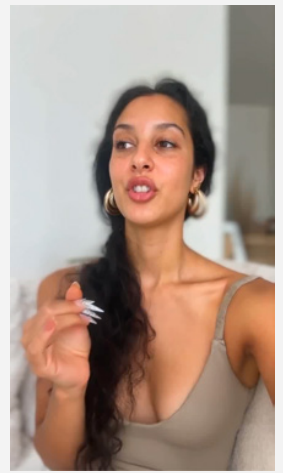

A young woman with a long, dark, messy braid cascading over her shoulder, her face adorned with subtle glitter that catches the light, and her lips slightly parted, exuding a relaxed yet captivating vibe. She engages with the camera in a serene manner, her expression inviting and thoughtful, as she occasionally tilts her head slightly, enhancing her minimal aesthetic. The background is softly blurred, featuring a minimalist indoor space with soft neutral tones, a few carefully placed decorative items, and gentle lighting that enhances the tranquil atmosphere. The shot is a medium, stationary frame at eye level, under bright, even lighting that creates a clear and engaging visual of the subject, highlighting her features and the simplicity of the setting.

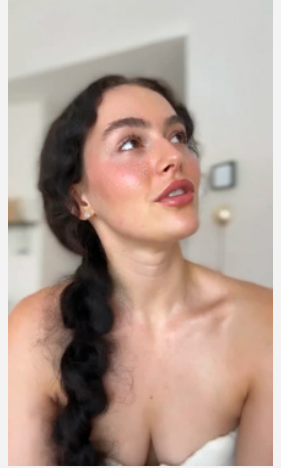

A middle-aged woman with a long, dark, messy braid and gentle features wears a plain white blouse, embodying a minimalistic style. She looks directly at the camera with her lips slightly parted, suggesting a moment of reflection and engagement with the viewer. The background is softly blurred, showcasing a simple indoor setting with light-colored walls, a few art pieces, and a cozy armchair, contributing to a calm and minimalist vibe. The shot is a medium, stationary angle at eye level, illuminated by bright, even lighting that creates a clear and engaging visual of her presence.

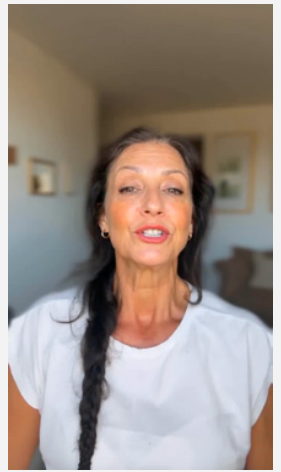

A young woman with a long, dark, messy braid adorned with a colorful silk scarf, her lips slightly parted, exuding a relaxed and minimal aesthetic. She engages softly with the camera, her expression calm and inviting, as she occasionally tilts her head slightly, enhancing her approachable demeanor. The background is softly blurred, featuring a minimalist indoor space with neutral-colored walls, a few carefully arranged decorative items, and natural light filtering through a nearby window, creating a serene atmosphere. The shot is a medium, stationary frame at eye level, under bright, even lighting that creates a clear and engaging visual of the subject while maintaining a polished and inviting composition.

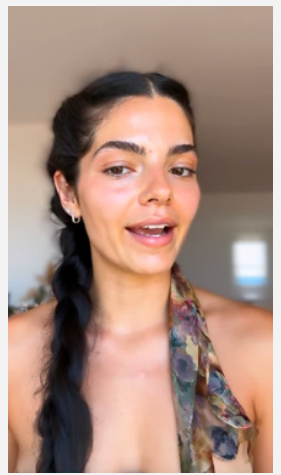

A young woman with a long, dark, messy braid that gives her a carefree vibe, complemented by large hoop earrings and striking long nails that reflect her personal style. She interacts with the camera, her lips slightly parted in a thoughtful expression, suggesting she is about to share something meaningful while exuding confidence. The background is softly blurred, showcasing a minimalistic indoor space with soft lighting, a few plants, and simple decor that creates a calm and inviting atmosphere. The shot is a medium, stationary frame at eye level, illuminated by bright, even lighting that enhances the clarity of her features and the overall composition.

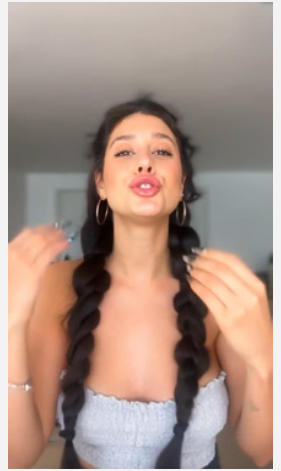

A young woman with a long, dark messy braid and a light dusting of freckles on her nose wears a relaxed greyish-blue hoodie, giving off a comfortable and approachable appearance. She looks thoughtfully at the camera, her lips slightly parted, suggesting a moment of contemplation as she prepares to share her thoughts. The background is softly blurred, showcasing a minimalistic indoor setting with soft lighting, a few decorative items on a shelf, and a calming color palette that promotes tranquility. The shot is a medium, stationary frame at eye level, illuminated by bright, even lighting that enhances the clarity and focus on her expressive features.

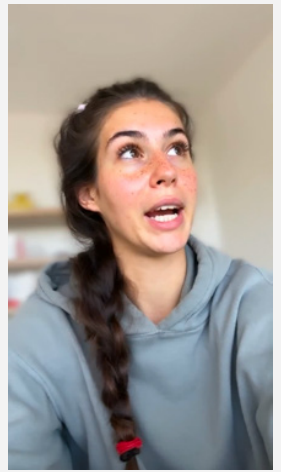

A young woman with a long, dark, messy braid draped over her shoulder wears a stylish white lacey blouse, which contrasts beautifully with her natural look. She engages gently with the camera, her lips slightly parted in a soft expression that invites connection and curiosity from the audience. The background is softly blurred, showcasing a minimalistic indoor space with light-colored walls, a few decorative items, and a warm ambiance that complements her outfit. The shot is a medium, stationary frame at eye level, illuminated by bright, even lighting that enhances the clarity of her features while creating an inviting visual experience.

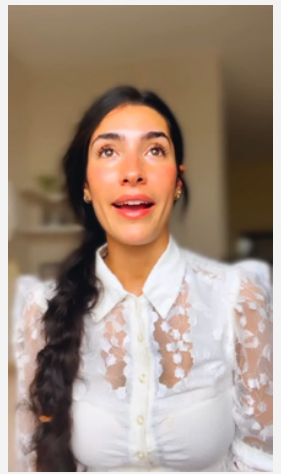

## 5.10. BACKGROUND AND PROP FIDELITY

Assesses environmental coherence, ensuring generated backgrounds (e.g., "sunlit café," "futuristic lab") and props (e.g., "holding a vintage microphone," "surrounded by holograms") match textual details without spatial or temporal inconsistencies. In general, Mirage performs the best with simple scenes, and tends to struggle when excessive complexity is added to the prompt. Each row of Fig. 17 displays a separate video, where each caption is the prompt used to generate the video.

● **Figure 17**  Select video frames and corresponding background prompts produced by Mirage.

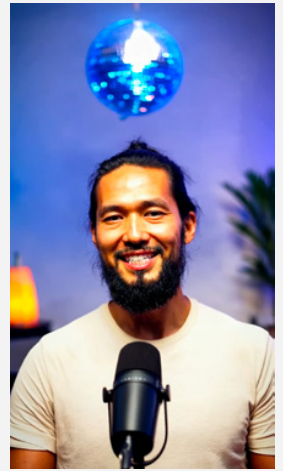

An Asian man with warm-toned skin, long black hair styled into a neat bun, and a full, well-groomed beard wears a casual off-white t-shirt, embodying a relaxed yet sophisticated vibe. He engages with the camera, offering a calm, assured smile that reflects his composed and approachable nature. The background is softly blurred, featuring a cozy interior with cool blue and purple ambient lighting, enhanced by a small orange-lit table lamp and a green potted plant that provide a touch of warmth. The shot is a medium close-up, stationary and well-lit, creating a serene and modern visual tone, with a disco ball slowly spinning above, casting gentle glimmers throughout the room.

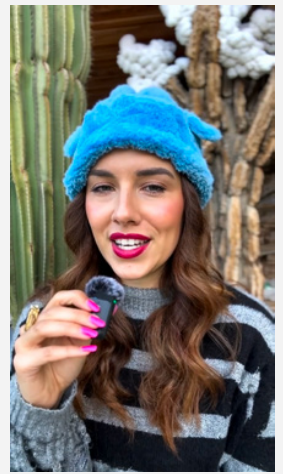

A young woman with fair skin and long, flowing brown hair, adorned with soft makeup that emphasizes her eyes and lips, wears a bright blue fuzzy hat with pointed ears, complemented by a black-and-gray striped sweater and glossy pink nail polish. She smiles brightly while speaking into a small gray microphone held close to her mouth, conveying a lively and animated presence that captivates the audience. The background is softly blurred, featuring a warm and inviting indoor space, prominently displaying a giant, gnarled saguaro cactus that adds an intriguing element to the cozy atmosphere. The shot is a medium close-up at eye level, illuminated by natural daylight that highlights her features and enhances the cheerful, playful mood of the scene.

An African-American man with a clean-shaven head, smooth dark skin, and a strong jawline is dressed in a charcoal gray button-up shirt with the top button unfastened, conveying a professional yet approachable look. He engages with the camera, maintaining steady eye contact while speaking in a calm and authoritative manner, his serious expression reflecting confidence and focus. The background is softly blurred, showcasing a series of computer monitors filled with cascading green digits and data, creating a visually captivating and high-tech environment. The shot is a medium close-up at eye level, stationary, under bright fluorescent lighting that clearly illuminates his face and enhances the vivid colors of the digital rain behind him.

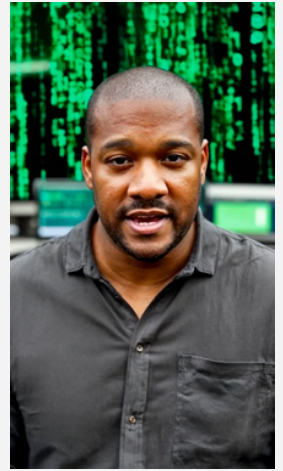

A young man with fair skin, shoulder-length blonde hair, and a light beard wears a fitted dark gray t-shirt that accentuates his athletic build. He looks directly into the camera, speaking animatedly with an engaged expression, his mouth slightly open mid-sentence while both hands are raised to emphasize his points. The background is softly blurred, showcasing a bustling gym filled with exercise equipment and several gymgoers, including multiple individuals clearly visible squatting at the squat racks. The shot is a medium close-up at eye level, captured in a stationary frame under harsh fluorescent lighting that creates a bright, clear visual of the subject.

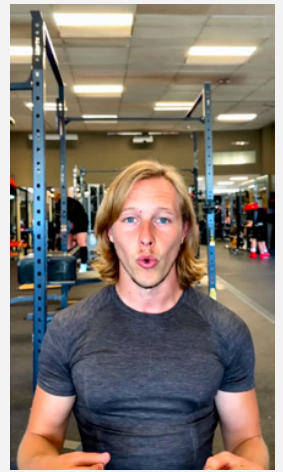

A woman with long, dark hair styled straight wears a brown ribbed sweater, complemented by gold hoop earrings and a nose ring, exuding a modern and fashionable vibe. She presents a misty crystal ball to the camera, her serious expression and direct engagement with a professional Shure microphone highlighting her focus and passion. The background is softly blurred, featuring a stylish, dimly lit setting with wooden slat paneling, warm ambient lighting, and a piano partially visible on the left, contributing to the sophisticated atmosphere. The shot is a medium close-up at eye level, illuminated by soft, directional lighting that creates subtle shadows, adding depth and a professional yet inviting tone.

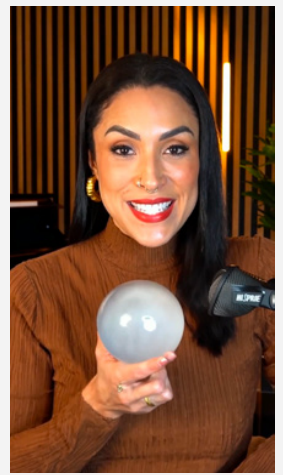

A young man with short, dark brown hair and round green-framed glasses, reflecting a hint of light, is dressed in a casual black shirt that complements his relaxed demeanor. He looks directly at the camera with a focused expression, slightly pursing his lips as he holds a large, vibrantly wrapped gift that captures the viewer's attention. The background is softly blurred, showcasing a cozy indoor environment with a gentle lamp glow and a decorative plant, contributing to a warm and inviting feel. The shot is a close-up at eye level with stationary framing, featuring warm, diffused lighting that enhances his skin tones and creates a polished, intimate atmosphere.

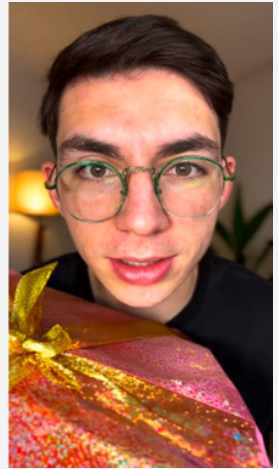

A young man with short, dark brown hair and round green-framed glasses, dressed in a stylish black shirt that highlights his engaged demeanor. He maintains eye contact with the camera, his lips slightly pursed as he speaks, prominently displaying a vibrant polka-dot cycling jersey in front of him. The background is softly blurred, featuring a cozy indoor scene with a lamp and a plant, adding to the inviting ambiance. The shot is a close-up at eye level with stationary framing, illuminated by warm, diffused lighting that enhances his features and creates a polished, intimate atmosphere.

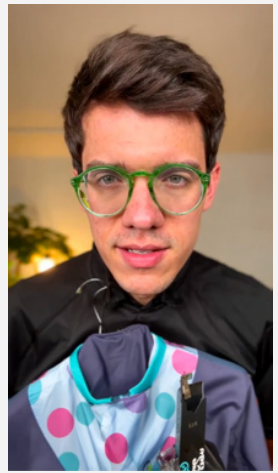

A young woman with long, dark hair styled straight wears a cozy brown ribbed sweater, adorned with gold hoop earrings and a nose ring, presenting a chic and modern look. She holds a stunning baseball glove towards the camera, her serious expression and direct engagement with a professional Shure microphone highlighting her focus and passion. The background is softly blurred, showcasing a stylish, dimly lit environment with wooden slat paneling, warm ambient lighting, and a piano partially visible on the left, contributing to an elegant ambiance. The shot is a medium close-up at eye level, featuring soft, directional lighting that creates subtle shadows, adding depth and a professional yet inviting feel.

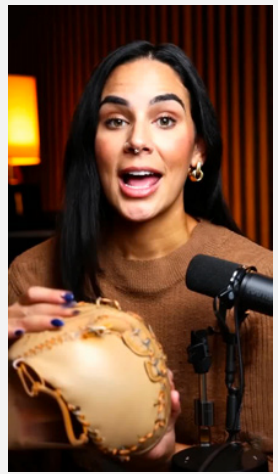

# CONCLUSION

In this report, we describe the Mirage audio-to-video generative model. Mirage is able to generate photorealistic, engaging performances with precisely synchronized audio and video when trained on A-roll footage. Mirage does this while making minimal modality-specific modifications to the underlying DiT architecture, allowing the core strategy to easily be extended to other conditional and multimodal generative settings.

Video (or *non-silent* video, in the common sense of the word) is multimodal, depending equally on imagery and audio to draw in viewers and tell stories convincingly. The goal of solving the broad problem of narrative video generation is to reliably create expressive, realistic, self-consistent, and compelling videos that faithfully and controllably capture all properties of visual and auditory modalities, as they would be deployed in expert-created film or video productions. Mirage introduces methodology that enables the generation of expressive visual performances from audio and in combination with other control signals, making a clear step towards the goal of fully flexible, realistic video storytelling. We encourage readers to watch and listen to the results of Mirage at [mirage.app/research/seeing-voices](mirage.app/research/seeing-voices) and to try Mirage for themselves at: [https://mirage.app](https://mirage.app).